\documentclass[10pt,twocolumn]{article}

\usepackage[letterpaper,
            top=0.75in,bottom=0.9in,
            left=0.72in,right=0.72in,
            columnsep=0.24in,
            headsep=0.2in]{geometry}

\usepackage[T1]{fontenc}
\usepackage{microtype}
\usepackage{balance}
\usepackage{helvet}

\usepackage{amsmath,amssymb}

\usepackage{graphicx}
\usepackage{booktabs}
\usepackage{multirow}
\usepackage{xcolor}
\usepackage{colortbl}
\usepackage{caption}
\usepackage{subcaption}
\usepackage{placeins}
\usepackage{dblfloatfix}
\usepackage{enumitem}
\usepackage{titlesec}
\usepackage{fancyhdr}
\usepackage{float}

\usepackage[colorlinks=true,
            linkcolor=odaxblue,
            citecolor=odaxblue!70!black,
            urlcolor=odaxblue]{hyperref}
\usepackage[capitalise,noabbrev]{cleveref}

\usepackage{algorithm}
\usepackage{algorithmic}

\definecolor{odaxblue}{RGB}{73,84,177}
\definecolor{odaxnavy}{RGB}{18,20,50}
\definecolor{odaxlight}{RGB}{247,247,253}

\titleformat{\section}
  {\fontsize{11.5}{13.5}\bfseries\color{odaxnavy}}
  {\thesection}{0.5em}{}
\titleformat{\subsection}
  {\fontsize{9.5}{11.5}\bfseries\color{odaxnavy}}
  {\thesubsection}{0.5em}{}
\titleformat{\subsubsection}
  {\fontsize{9}{11}\itshape\color{odaxnavy}}
  {\thesubsubsection}{0.5em}{}
\titlespacing{\section}{0pt}{10pt plus 2pt}{3pt}
\titlespacing{\subsection}{0pt}{7pt plus 1pt}{2pt}
\titlespacing{\subsubsection}{0pt}{5pt}{1pt}

\setlist[itemize]{leftmargin=*,topsep=2pt,itemsep=1pt,parsep=0pt}

\newcommand{\vH}{\textsf{Venice-H1}}
\DeclareMathOperator*{\argmax}{arg\,max}
\newcommand{\ie}{i.e.}
\newcommand{\eg}{e.g.}
\makeatletter
\long\gdef\@myabstract{}
\newcommand{\setabstract}[1]{\long\gdef\@myabstract{#1}}

\renewcommand{\maketitle}{%
  \twocolumn[{%
    \vspace*{-4pt}
    \begin{center}
      \raisebox{-6pt}{\includegraphics[height=34pt]{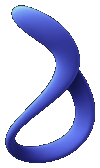}}%
      \hspace{6pt}%
      {\usefont{T1}{lato-LF}{m}{n}\fontsize{26}{26}\selectfont\color{odaxnavy}OdaxAI}
    \end{center}
    \vspace{14pt}
    \begin{center}
      {\fontsize{16.5}{20}\bfseries\color{odaxnavy} \@title \par}
      \vspace{10pt}
      {\usefont{T1}{lato-LF}{m}{n}\fontsize{11.5}{14}\selectfont\color{odaxnavy}%
        Nicol\`o Savioli,\,{\fontsize{10.5}{12}\selectfont Ph.D.}\par}
      \vspace{4pt}
      {\usefont{T1}{lato-LF}{m}{n}\fontsize{13.5}{16}\selectfont%
        \href{https://odaxai.com}{\color{odaxblue}\textls[80]{OdaxAI Research}}\par}
      \vspace{6pt}
      {\fontsize{8}{10}\selectfont\color{gray!65}%
        \texttt{nicolo.savioli@odaxai.com}%
        \hspace{8pt}{\color{gray!40}$\vert$}\hspace{8pt}%
        \href{https://odaxai.com}{\color{odaxblue!70}www.odaxai.com}%
      \par}
      \vspace{8pt}
      {\fontsize{7.8}{10}\selectfont%
        \colorbox{gray!8}{%
          \hspace{6pt}%
          \href{https://github.com/odaxai/Venice-H1}{%
            \raisebox{-1pt}{\includegraphics[height=9pt]{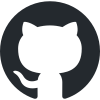}}%
            \hspace{3pt}\color{odaxnavy}\textsf{Code}}%
          \hspace{14pt}%
          \href{https://huggingface.co/OdaxAI/venice-h1}{%
            \raisebox{-1.5pt}{\includegraphics[height=10pt]{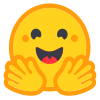}}%
            \hspace{3pt}\color{odaxnavy}\textsf{Model}}%
          \hspace{6pt}%
        }%
      \par}
    \end{center}
    \vspace{12pt}
    \noindent\hspace{0.04\linewidth}%
    \fcolorbox{odaxblue!15}{odaxlight}{%
      \begin{minipage}{0.895\linewidth}
        \vspace{6pt}
        \fontsize{8.5}{11.5}\selectfont
        \setlength{\parskip}{3pt}\setlength{\parindent}{0pt}%
        \@myabstract
        \vspace{4pt}
      \end{minipage}%
    }\par
    \vspace{14pt}
  }]%
  \setcounter{footnote}{0}%
}
\makeatother

\fancypagestyle{plain}{\fancyhf{}\fancyfoot[C]{\fontsize{7.5}{8}\selectfont\color{gray!70}\thepage}\fancyfoot[R]{\fontsize{6.5}{8}\selectfont\color{gray!50}\href{https://arxiv.org/abs/2606.22546}{arXiv:2606.22546}}}

\begin{document}

\title{Venice-H1: Failure-Aware Query Re-Ranking with\\[2pt]
Multi-Scale Grid Signatures for Referring Image Segmentation}

\setabstract{%
Modern Referring Image Segmentation (RIS) systems generate multiple candidate masks per expression but
rely on a simple heuristic---typically the argmax detection score---to select the final output.
We identify \emph{query selection} as a failure-case bottleneck: although heuristic selection succeeds on
82--93\% of samples, the residual 7--18\% of failures dominate the error budget, leaving a best-query
selection gap of 3--11\% mIoU.
We introduce \vH{}, a lightweight, backbone-decoupled post-hoc re-ranking module that encodes each
candidate through \emph{multi-scale grid signatures}---compact spatial descriptors pooled onto
$4{\times}4$, $8{\times}8$, and $16{\times}16$ grids---and feeds them to a Transformer-based re-ranker
with a \emph{Failure Gate} (ROC-AUC 0.78--0.82) that intervenes only when the default choice is likely
suboptimal.
Instantiated on DeRIS-L and DeRIS-B, \vH{} achieves $\Delta_{\mathrm{fail}}$ of +1.40 and +0.89 mIoU
with strictly positive 95\% CIs on all 16/16 (split, backbone) pairs and harmful-switch rates below 0.53\%.
Zero-shot transfer to medical referring segmentation (MS-CXR, M3D-RefSeg-2D) yields +1.16 and
+0.51 mIoU without RIS-backbone fine-tuning.
The module adds ${\sim}$11.3\,M parameters and $<$1\,ms latency.%
}

\maketitle

\section{Introduction}
\label{sec:intro}

Referring Image Segmentation (RIS)~\cite{refcoco,refcocog} requires predicting a pixel-accurate mask for the image region described by a free-form natural language expression.
Modern architectures~\cite{lavt,cris,oneref,deris} achieve remarkable performance through large-scale vision--language pretraining~\cite{beit3} and sophisticated multi-query prediction pipelines.
In particular, decoupled frameworks such as DeRIS~\cite{deris} generate $N$ candidate query masks per sample and select the top-ranked one (Query\,0) as the final prediction.
This multi-hypothesis design is not specific to DeRIS: it arises in any model with a set of learned queries (e.g., Mask2Former~\cite{mask2former}), making the query-selection problem a general bottleneck for multi-query RIS architectures.

We begin with a diagnostic observation (\Cref{fig:error_budget}): across eight evaluation splits of RefCOCO, RefCOCO+, and RefCOCOg, DeRIS-L's default query selection is suboptimal in 7--18\% of samples (we term these \emph{failure cases}: instances where $\argmax_i \mathrm{IoU}(\mathbf{P}_i,\mathbf{y}) \neq \argmax_i s_i$).
Although these failures are a minority, they contribute disproportionately to the overall error: 40--68\% of the total segmentation error budget originates from this subset, with best-query mIoU gaps of 3--11\%.
The correct mask \emph{exists} among the candidates in every failure case---it is simply not selected by the default heuristic.
This raises a precise question: can we detect when the default query fails and select a better alternative, without modifying the base model?

The challenge lies in building discriminative features that distinguish good from bad query masks.
Raw mask logits are high-dimensional and vary in resolution; directly comparing them is expensive and noisy.
We propose \emph{multi-scale grid signatures}: compact spatial descriptors obtained by pooling mask probabilities onto structured grids at three resolutions ($4{\times}4$, $8{\times}8$, $16{\times}16$).
At each scale, we extract the grid-cell mean (spatial coverage), grid-cell max (peak activation), and a boundary-energy scalar that quantifies edge strength, yielding a 675-dimensional representation per candidate.
Coarse grids encode global layout; fine grids capture local shape and boundary quality.
The design is inspired by multi-scale grid-cell representations in the mammalian entorhinal cortex~\cite{gridcells,hafting2005microstructure}, where overlapping periodic codes at different spatial frequencies support robust spatial encoding~\cite{banino2018}.

Using grid signatures, we construct \vH{}: a lightweight, backbone-decoupled re-ranking module that operates entirely on cached features extracted from a frozen backbone.
A compact Transformer-based re-ranker processes the per-query feature vectors through two heads: a \emph{Failure Gate} that predicts whether Query\,0 is suboptimal ($\hat{p}_{\mathrm{fail}}$), and a \emph{Gain Predictor} that estimates the IoU improvement $\hat{g}_i$ each alternative query would provide.
At inference, if $\hat{p}_{\mathrm{fail}} > \tau$, the system selects $\argmax_i \hat{g}_i$; otherwise it retains Query\,0.
This conservative gated design intervenes only on predicted failures, preserving baseline accuracy on the 82--93\% of samples that are already optimally selected.

We instantiate \vH{} on DeRIS-L and DeRIS-B and evaluate across all eight standard RefCOCO/+/g splits.
Our main contributions are:
\begin{itemize}
  \item We identify and quantify the \emph{failure-case bottleneck} in multi-query RIS: 7--18\% of evaluation samples concentrate 40--68\% of the total error budget, with recoverable gaps up to 11\% mIoU (\Cref{sec:diagnostic}).
  \item We propose compact 675-dimensional \emph{multi-scale grid signatures} at three resolutions that encode mask shape, coverage, and boundary quality. Applied on a frozen C3VG backbone, they yield +4.17\% mIoU alone, confirming backbone-decoupled discriminability (\Cref{sec:generalizability}).
  \item We propose \vH{}, a gated Transformer re-ranker achieving gate AUC 0.78--0.82, strictly positive $\Delta_{\mathrm{fail}}$ on 16/16 (split, backbone) pairs, and $<$1\,ms overhead, with no modification to the base model.
  \item Without any fine-tuning, \vH{} transfers to medical referring segmentation (MS-CXR, M3D-RefSeg-2D), recovering 12--14\% of the best-query gap despite severe domain shift (\Cref{sec:medical}).
\end{itemize}

\FloatBarrier
\section{Related Work}
\label{sec:related}

RIS methods fuse vision and language to predict pixel-accurate masks from free-form expressions.
Perception-centric approaches modulate visual features with language via post-fusion~\cite{cmpc,lscm} or early-fusion~\cite{lavt,efn}.
Cognition-centric methods leverage vision--language pretraining~\cite{clip,beit3} for richer cross-modal alignment~\cite{cris,oneref,c3vg}.
State-of-the-art specialist methods such as DeRIS~\cite{deris} decouple perception (Swin~\cite{swinb}+Mask2Former~\cite{mask2former}) from cognition (BEiT-3~\cite{beit3}), generating $N$ candidate masks and selecting the top-ranked one.
Our work is complementary: we operate after candidate generation and focus exclusively on improving query selection.

Generating multiple candidates and selecting the best is widespread in segmentation and detection.
SAM~\cite{sam} outputs multiple masks and uses a learned IoU head for ranking; Mask Scoring R-CNN~\cite{maskscoring} regresses mask IoU to recalibrate detection scores; IoU-Net~\cite{iounet} formalizes the gap between classification and spatial accuracy.
In weakly supervised RIS, Segment-Select-Correct~\cite{ssc} explicitly decomposes the pipeline into generation, selection, and refinement.
Unlike these per-mask quality estimators (one mask in, one score out), \vH{} jointly models all $K$ candidates with an explicit, IoU-supervised gate sub-task, treating re-ranking as a set-level prediction problem.
Plug-and-play post-processing has a long history: Soft-NMS~\cite{softnms} and Learning NMS~\cite{learningnms} improve detection without retraining.
Our Failure Gate draws on selective prediction~\cite{geifman2017selective} and SelectiveNet~\cite{selectivenet}, which learn coverage--risk trade-offs, but uses explicit binary supervision and rich spatial features rather than logit uncertainty alone.

Spatial pyramid pooling~\cite{sppnet} established grid-based pooling for spatial layout encoding.
Grid-like codes originate from neuroscience: entorhinal grid cells provide periodic spatial representations at multiple frequencies~\cite{gridcells,hafting2005microstructure}, inspiring navigation agents~\cite{banino2018} and spatial encoding models~\cite{mai2020,schaeffer2023}.
We transfer this principle to RIS, pooling mask probabilities onto $4{\times}4$, $8{\times}8$, and $16{\times}16$ grids to build 675-dimensional signatures that capture complementary spatial scales.
For training, focal loss~\cite{lin2017focal} addresses class imbalance; we use IoU regression as a pointwise surrogate providing dense supervision from all samples, circumventing data starvation on the 0.28\% failure subset where binary cross-entropy collapses.

\FloatBarrier
\section{Method}
\label{sec:method}

We present \vH{}, a post-hoc failure-aware re-ranking framework for multi-query Referring Image Segmentation.
The key idea is to detect when the default query selection (Query\,0) is suboptimal and, when so, select the alternative query with the highest predicted IoU gain.

\paragraph{Interface requirements.}
\vH{} requires only three outputs from the upstream model: (i)~a set of $N$ candidate mask predictions, (ii)~per-candidate confidence scores, and (iii)~optional query embeddings.
It does not access backbone internals, loss functions, or training-time signals, and plugs in after candidate generation with $O(N)$ complexity per sample.
We use DeRIS-L as a concrete instantiation, but the interface is satisfied by any multi-hypothesis RIS system.

\begin{figure*}[t]
  \centering
  \includegraphics[width=\linewidth]{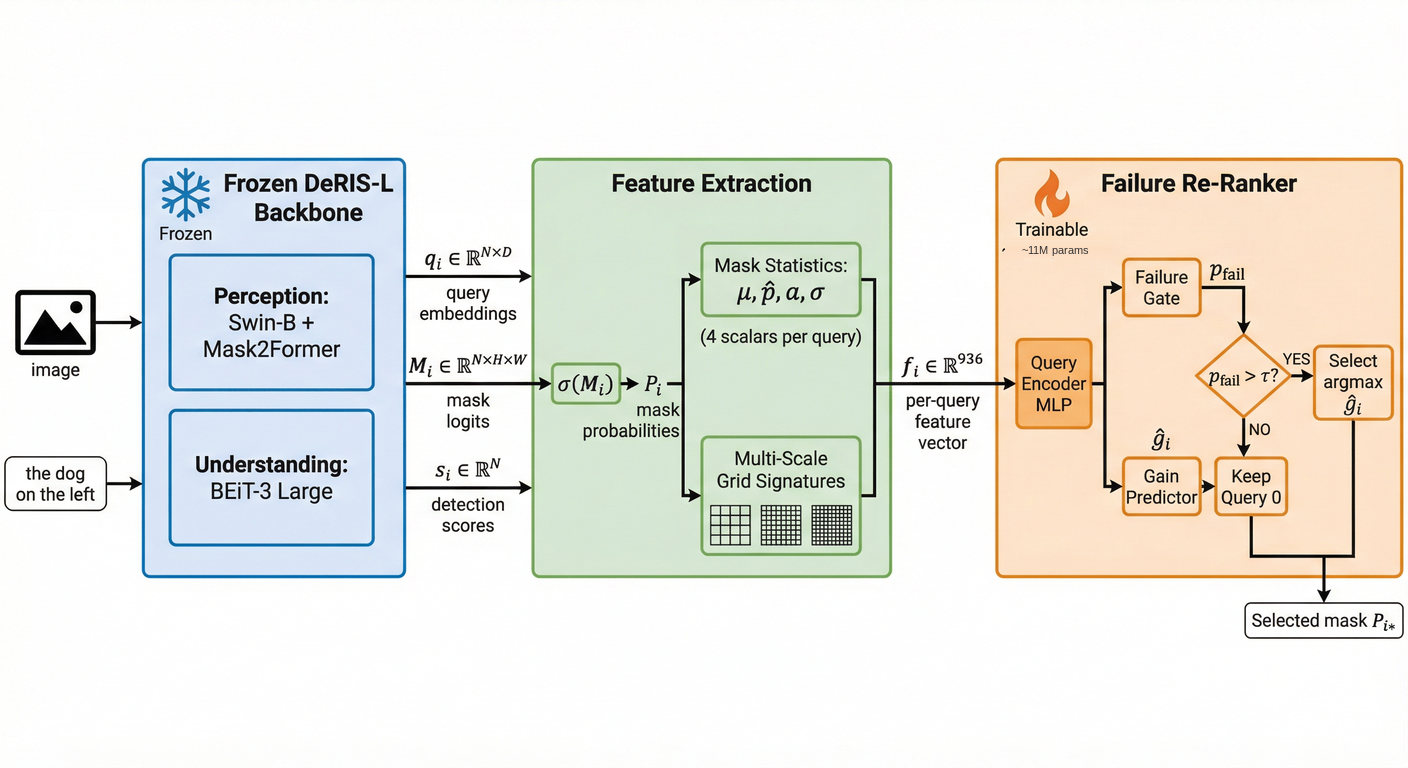}
  \caption{\textbf{\vH{} pipeline overview.}
  A frozen DeRIS-L backbone (left, blue) produces $N$ query embeddings $\mathbf{q}_i$, mask logit maps $\mathbf{M}_i$, and detection scores $s_i$.
  The feature extraction stage (center) computes mask statistics and multi-scale grid signatures from mask probabilities $\mathbf{P}_i$.
  The Failure Re-Ranker (right, orange) uses a Transformer-based architecture with two heads: a Failure Gate predicting $\hat{p}_{\mathrm{fail}}$ and a Gain Predictor estimating $\hat{g}_i$ for each query.
  If $\hat{p}_{\mathrm{fail}} > \tau$, the system selects $\argmax_i \hat{g}_i$; otherwise Query\,0 is retained.
  Only the re-ranker parameters (${\sim}11$\,M) are trained.}
  \label{fig:architecture}
\end{figure*}

\Cref{fig:architecture} illustrates the full pipeline.
We describe the frozen backbone and feature extraction (\Cref{sec:backbone,sec:features}), the multi-scale grid signature design (\Cref{sec:gridsig}), the re-ranking model (\Cref{sec:reranker}), and the training procedure (\Cref{sec:training}).

\subsection{Frozen Backbone}
\label{sec:backbone}

We build on DeRIS-L~\cite{deris}, which comprises a Perception branch (Swin-B~\cite{swinb} + Mask2Former~\cite{mask2former}) and an Understanding branch (BEiT-3 Large~\cite{beit3}).
Given an image $I$ and referring expression $e$, DeRIS produces $N$ candidate predictions:
\begin{itemize}
  \item Query embeddings $\{\mathbf{q}_i\}_{i=0}^{N-1} \subset \mathbb{R}^{D}$, encoding semantic--spatial information for each candidate.
  \item Mask logit maps $\{\mathbf{M}_i\}_{i=0}^{N-1}$, where $\mathbf{M}_i \in \mathbb{R}^{H \times W}$.
  \item Detection scores $\{s_i\}_{i=0}^{N-1} \subset [0,1]$, ranking candidates by relevance.
\end{itemize}
The default prediction is Query\,0, \ie, $i^*_{\mathrm{base}} = 0$.
All backbone parameters are frozen; \vH{} operates entirely on the extracted outputs.

\subsection{Feature Extraction}
\label{sec:features}

For each sample, we compute mask probabilities by applying a sigmoid to the logits:
\begin{equation}
  \mathbf{P}_i = \sigma(\mathbf{M}_i) \;\in\; [0,1]^{H \times W}, \quad i = 0, \ldots, N{-}1.
  \label{eq:mask_probs}
\end{equation}
From these, we extract four scalar \emph{mask statistics} per query: mean coverage $\mu_i{=}\mathrm{mean}(\mathbf{P}_i)$, peak confidence $\hat{p}_i{=}\max(\mathbf{P}_i)$, binary area $a_i{=}\mathrm{mean}(\mathbb{1}[\mathbf{P}_i{>}0.5])$, and spatial uncertainty $\sigma_i{=}\mathrm{std}(\mathbf{P}_i)$.

\subsection{Multi-Scale Grid Signatures}
\label{sec:gridsig}

Scalar mask statistics capture global properties but discard spatial structure.
To encode \emph{where} a mask is active---and distinguish, \eg, a correctly localized mask from one that covers the wrong region---we introduce \emph{multi-scale grid signatures}.

For each query $i$ and grid resolution $G \in \{4, 8, 16\}$, we pool $\mathbf{P}_i$ onto a $G{\times}G$ lattice:
\begin{align}
  \mathbf{g}^{\mu}_{i,G}   &= \mathrm{AvgPool}_{G{\times}G}(\mathbf{P}_i)
    \;\in\; \mathbb{R}^{G^2},
  \label{eq:grid_mean} \\
  \mathbf{g}^{\max}_{i,G}  &= \mathrm{MaxPool}_{G{\times}G}(\mathbf{P}_i)
    \;\in\; \mathbb{R}^{G^2},
  \label{eq:grid_max}
\end{align}
where $\mathbf{g}^{\mu}_{i,G}$ captures the average activation per cell (coverage pattern) and $\mathbf{g}^{\max}_{i,G}$ captures the peak activation (presence of high-confidence regions).
Both use adaptive pooling to handle variable input resolutions.

We additionally extract a scalar \emph{boundary energy} that quantifies the spatial gradient magnitude across the grid.
Let $\nabla_x$ and $\nabla_y$ denote the horizontal and vertical finite differences on the $G{\times}G$ grid:
\begin{equation}
  b_{i,G}
  = \tfrac{1}{2}\bigl(\,
    \overline{|\nabla_x\, \mathbf{g}^{\mu}_{i,G}|}
    \;+\;
    \overline{|\nabla_y\, \mathbf{g}^{\mu}_{i,G}|}
  \,\bigr),
  \label{eq:boundary}
\end{equation}
where $\overline{|\cdot|}$ denotes the mean absolute value over all adjacent cell pairs.
This is the average absolute gradient of the mean-pooled grid.
High boundary energy indicates well-defined mask edges; low values suggest a diffuse or empty mask.

\paragraph{Multi-scale complementarity.}
The coarse grid ($4{\times}4$) encodes global layout (which quadrant is active); the medium grid ($8{\times}8$) captures intermediate shape; the fine grid ($16{\times}16$) encodes boundary detail and local structure.

\paragraph{Grid signature vector.}
The full grid signature for query $i$ concatenates all three scales:
\begin{equation}
  \mathbf{g}_i = \bigl[\,
    \mathbf{g}^{\mu}_{i,4};\,\mathbf{g}^{\max}_{i,4};\,b_{i,4};\;\ldots;\;
    \mathbf{g}^{\mu}_{i,16};\,\mathbf{g}^{\max}_{i,16};\,b_{i,16}
  \,\bigr],
  \label{eq:grid_concat}
\end{equation}
yielding $\mathbf{g}_i \in \mathbb{R}^{675}$, since each scale contributes $(G^2 + G^2 + 1)$ values: $33 + 129 + 513 = 675$.

\begin{figure*}[t]
\centering
\includegraphics[width=\linewidth]{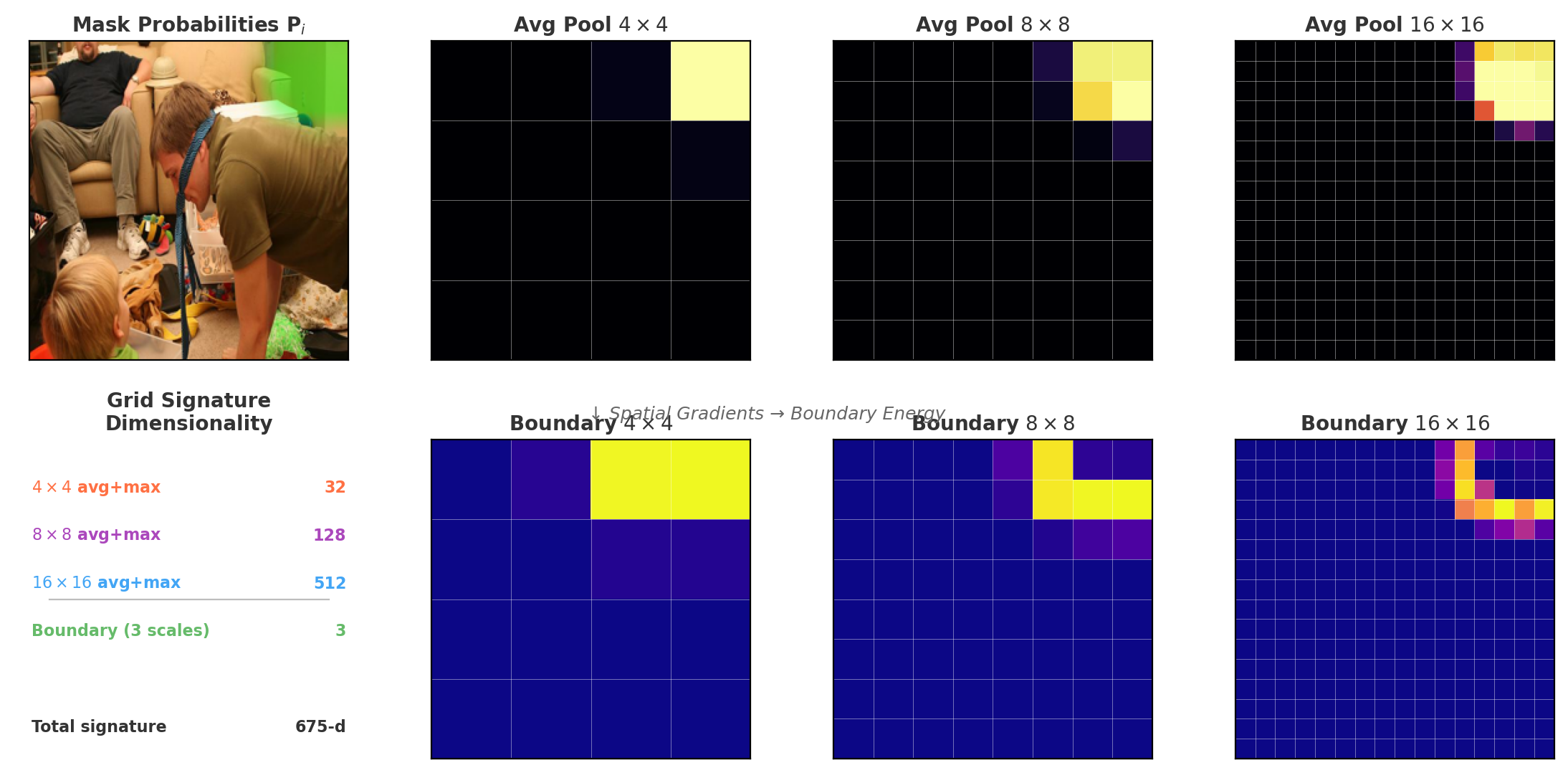}
\caption{\textbf{Multi-scale grid signatures on a RefCOCO example.}
From the mask probability map $\mathbf{P}_i$, we compute grid signatures at $4{\times}4$ (coarse), $8{\times}8$ (medium), and $16{\times}16$ (fine) resolutions.
Each scale captures different spatial aspects: location, shape, and boundary detail.
The combined 675-dim signature provides a compact yet discriminative representation for re-ranking.}
\label{fig:gridcells_supp}
\end{figure*}

\Cref{fig:gridcells_supp} illustrates how multi-scale grid signatures encode mask spatial structure on a concrete example.
The combined 675-dimensional signature captures complementary information at three resolutions: coarse location ($4{\times}4$), shape ($8{\times}8$), and fine boundary detail ($16{\times}16$).

\subsection{Failure Re-Ranker}
\label{sec:reranker}

The re-ranking model comprises a query encoder, a Transformer encoder for inter-query reasoning, and two task-specific heads.

\paragraph{Per-query feature assembly.}
For each query $i$, we concatenate the query embedding, detection score, mask statistics, and grid signature:
\begin{equation}
  \mathbf{f}_i = \big[\,\mathbf{q}_i;\; s_i;\; \mu_i;\; \hat{p}_i;\; a_i;\; \sigma_i;\; \mathbf{g}_i\,\big]
  \;\in\; \mathbb{R}^{D_f},
  \label{eq:per_query_feat}
\end{equation}
where $D_f = D + 1 + 4 + 675 = D + 680$.
For $D{=}256$ (DeRIS query dimension), $D_f{=}936$.
Without grid signatures (\ie, the BASE feature set), $D_f{=}261$.

\paragraph{Query encoder and Transformer.}
Each $\mathbf{f}_i$ is projected into $\mathbb{R}^{H_d}$ ($H_d{=}512$) by a shared two-layer MLP with LayerNorm and ReLU, then processed by an $L$-layer pre-norm Transformer encoder ($L{=}3$, $A{=}8$ heads, GELU FFN):
\begin{equation}
  \mathbf{h}_i^{(\ell)}
  = \mathrm{TransformerLayer}^{(\ell)}\!\bigl(\mathbf{h}_i^{(\ell-1)}\bigr),
  \quad \ell = 1,\ldots,L.
  \label{eq:transformer}
\end{equation}
Self-attention enables inter-query comparison---detecting outlier masks that deviate from the consensus---a signal not accessible to per-query features alone.
We denote $\mathbf{h}_i \equiv \mathbf{h}_i^{(L)}$.

\paragraph{Head 1: Gain Predictor.}
A two-layer MLP predicts the IoU improvement of each query over the default:
\begin{equation}
  \hat{g}_i = \mathbf{w}_g^{\!\top}\, \mathrm{ReLU}(\mathbf{W}_g\, \mathbf{h}_i + \mathbf{b}_g) + c_g
  \;\in\; \mathbb{R},
  \label{eq:gain_pred}
\end{equation}
where the target is $g_i^* = \mathrm{IoU}(\mathbf{P}_i, \mathbf{y}) - \mathrm{IoU}(\mathbf{P}_0, \mathbf{y})$ and $\mathbf{y}$ is the ground-truth mask.
Positive values indicate that query $i$ outperforms the default.

\paragraph{Head 2: Failure Gate.}
A separate MLP predicts whether the default selection is suboptimal from a global feature vector $\mathbf{r} = [\bar{\mathbf{h}};\;\hat{\mathbf{h}};\;\mathbf{h}_{\mathrm{def}};\;\mathbf{s};\;\boldsymbol{\mu}] \in \mathbb{R}^{3H_d + 2N}$, where $\bar{\mathbf{h}}$, $\hat{\mathbf{h}}$ are mean/max-pooled Transformer outputs, $\mathbf{h}_{\mathrm{def}}$ is the default query's output, and $\mathbf{s}$, $\boldsymbol{\mu}$ are raw scores and mask means:
\begin{equation}
  \hat{p}_{\mathrm{fail}} = \sigma\!\big(\mathrm{MLP}_f(\mathbf{r})\big).
  \label{eq:failure_gate}
\end{equation}

\paragraph{Gated query selection.}
At inference, the re-ranker selects the final query via:
\begin{equation}
  i^* = \begin{cases}
    \argmax_{i}\; \hat{g}_i & \text{if } \hat{p}_{\mathrm{fail}} > \tau, \\[3pt]
    0 & \text{otherwise},
  \end{cases}
  \label{eq:gated_selection}
\end{equation}
where $\tau \in (0,1)$ is a threshold tuned on the validation set.
The gated design ensures that the re-ranker intervenes only when confident that Query\,0 is wrong, preserving the strong baseline accuracy on the majority of samples.

\subsection{Training}
\label{sec:training}

\vH{} is trained entirely on cached features using a multi-task loss.

\paragraph{Loss function.}
The total loss $\mathcal{L} = \mathcal{L}_{\mathrm{gate}} + \lambda\, \mathcal{L}_{\mathrm{gain}}$ ($\lambda{=}5$) combines:
(i)~focal BCE~\cite{lin2017focal} ($\gamma{=}2$, auto $w_{\mathrm{pos}}$) on the binary failure label $y_{\mathrm{fail}} = \mathbb{1}[\argmax_i \mathrm{IoU}(\mathbf{P}_i, \mathbf{y}) \neq 0]$; and
(ii)~smooth-$\ell_1$ IoU regression on \emph{all} samples:
\begin{equation}
  \mathcal{L}_{\mathrm{gain}} = \tfrac{1}{BN}
    \textstyle\sum_{b,i} \mathrm{SL1}\!\bigl(\hat{g}_i^{(b)},\,
    \Delta\mathrm{IoU}_i^{(b)}\bigr),
  \label{eq:gain_loss}
\end{equation}
where $\Delta\mathrm{IoU}_i = \mathrm{IoU}(\mathbf{P}_i, \mathbf{y}) - \mathrm{IoU}(\mathbf{P}_0, \mathbf{y})$.
IoU regression on all 126\,k samples provides dense supervision, unlike cross-entropy which starves on the 0.28\% failure subset.

\paragraph{Optimization.}
AdamW with cosine annealing, 20 epochs on cached features (batch 512, FP16).
The threshold $\tau$ is tuned via grid search on the validation set, maximizing $\Delta_{\mathrm{full}}$.

\paragraph{Inference.}
At test time: (1)~DeRIS-L forward pass, (2)~grid signature extraction (three adaptive pooling ops per query), (3)~re-ranker forward pass.
Steps (2)--(3) add $<1$\,ms on GPU.
The full inference procedure is given in \Cref{alg:inference}.

\begin{algorithm}[t]
\caption{\vH{} Inference}
\label{alg:inference}
\begin{algorithmic}[1]
\REQUIRE Image $I$, expression $e$, threshold $\tau$
\STATE $\{\mathbf{q}_i, \mathbf{M}_i, s_i\}_{i=0}^{N-1} \leftarrow \mathrm{DeRIS}(I, e)$ \hfill\COMMENT{frozen backbone}
\STATE $\mathbf{P}_i \leftarrow \sigma(\mathbf{M}_i)$ for all $i$ \hfill\COMMENT{mask probabilities}
\STATE $\{\mu_i, \hat{p}_i, a_i, \sigma_i\} \leftarrow$ mask statistics from $\mathbf{P}_i$
\STATE $\mathbf{g}_i \leftarrow$ grid signatures from $\mathbf{P}_i$ at scales $\{4,8,16\}$
\STATE $\mathbf{f}_i \leftarrow [\mathbf{q}_i; s_i; \mu_i; \hat{p}_i; a_i; \sigma_i; \mathbf{g}_i]$
\STATE $\hat{p}_{\mathrm{fail}}, \{\hat{g}_i\} \leftarrow \mathrm{ReRanker}(\{\mathbf{f}_i\})$
\IF{$\hat{p}_{\mathrm{fail}} > \tau$}
  \STATE $i^* \leftarrow \argmax_i \hat{g}_i$
\ELSE
  \STATE $i^* \leftarrow 0$
\ENDIF
\RETURN mask $\mathbf{P}_{i^*}$
\end{algorithmic}
\end{algorithm}

\FloatBarrier
\section{Experiments}
\label{sec:experiments}

\subsection{Datasets and Metrics}
\label{sec:datasets}

We evaluate on three standard RIS benchmarks.
\textbf{RefCOCO}~\cite{refcoco} (142\,k expressions, 50\,k objects) is split into val, testA (person-centric), and testB (non-person).
\textbf{RefCOCO+}~\cite{refcoco} prohibits absolute location words, stressing appearance reasoning.
\textbf{RefCOCOg}~\cite{refcocog} features longer expressions (avg.\ 8.4 words) under the UMD split.
We report mean IoU (mIoU) following standard protocol.

\subsection{Implementation Details}
\label{sec:implementation}

\paragraph{Backbone and features.}
We instantiate \vH{} on DeRIS-L~\cite{deris}, the current specialist SOTA.
The Perception branch (Swin-B~\cite{swinb} + Mask2\-Former~\cite{mask2former}) and Understanding branch (BEiT-3 Large~\cite{beit3}) are kept entirely frozen.
Features are extracted offline: query embeddings ($N{=}10$, $D{=}256$), detection scores, mask statistics, and multi-scale grid signatures.
Total cached features: ${\sim}$3.2\,GB (126\,k training samples).

\paragraph{Re-ranker.}
Feature set: BASE+GRID ($D_f{=}936$ per query), hidden dim $H_d{=}512$.
3-layer Transformer encoder, 8 heads, pre-norm GELU, 10\% dropout.
Total: ${\sim}$11.3\,M trainable parameters.

\paragraph{Optimization.}
AdamW ($\mathrm{lr}{=}5{\times}10^{-4}$, wd $10^{-4}$), cosine schedule with 3-epoch warmup, 20 epochs, batch 512, FP16.
Training set: 126\,k samples (0.28\% failure rate, ${\approx}$354 positives).
This low rate is expected: the training split was ``seen'' by DeRIS during its own training, so Q0 is nearly always optimal.
At evaluation time, distribution shift raises the failure rate to 7--18\% (see \Cref{sec:diagnostic}).
We compensate the class imbalance via IoU regression on \emph{all} samples (rather than binary classification on the rare positives) and focal BCE ($\gamma{=}2$) with automatic positive weight $w_{\mathrm{pos}}$.
Training: ${\sim}$3\,min on a single GPU.

\subsection{Diagnostic Analysis: The Failure-Case Bottleneck}
\label{sec:diagnostic}

\Cref{tab:diagnostic} reports the gap between DeRIS-L's default selection and the best-query upper bound.

\begin{table}[t]
\centering
\caption{\textbf{Query selection gap analysis} (evaluated in our pipeline).
Default: top-scoring query (Q0). Best-query: best among $N{=}10$ candidates per sample.
The best-query upper bound reveals 3--11\% mIoU room for post-hoc re-ranking.
Note: absolute mIoU values differ slightly from Table~\ref{tab:sota} ($\leq$0.4\%), as IoU is computed at model output resolution; \textit{gaps} are resolution-independent.}
\label{tab:diagnostic}
\resizebox{\columnwidth}{!}{%
\begin{tabular}{l|ccc|ccc|cc}
\toprule
 & \multicolumn{3}{c|}{RefCOCO} & \multicolumn{3}{c|}{RefCOCO+} & \multicolumn{2}{c}{RefCOCOg} \\
 & val & testA & testB & val & testA & testB & val & test \\
\midrule
Default (Q0)    & 85.46 & 86.47 & 84.57 & 81.12 & 83.38 & 78.63 & 79.80 & 81.11 \\
Best-query (UB) & 89.43 & 89.69 & 89.70 & 89.29 & 89.49 & 89.75 & 87.28 & 88.00 \\
\rowcolor{gray!10}
Gap             & +3.97 & +3.22 & +5.13 & +8.16 & +6.12 & \textbf{+11.11} & +7.48 & +6.89 \\
\bottomrule
\end{tabular}}
\end{table}

\begin{figure}[t!]
\centering
\includegraphics[width=\columnwidth]{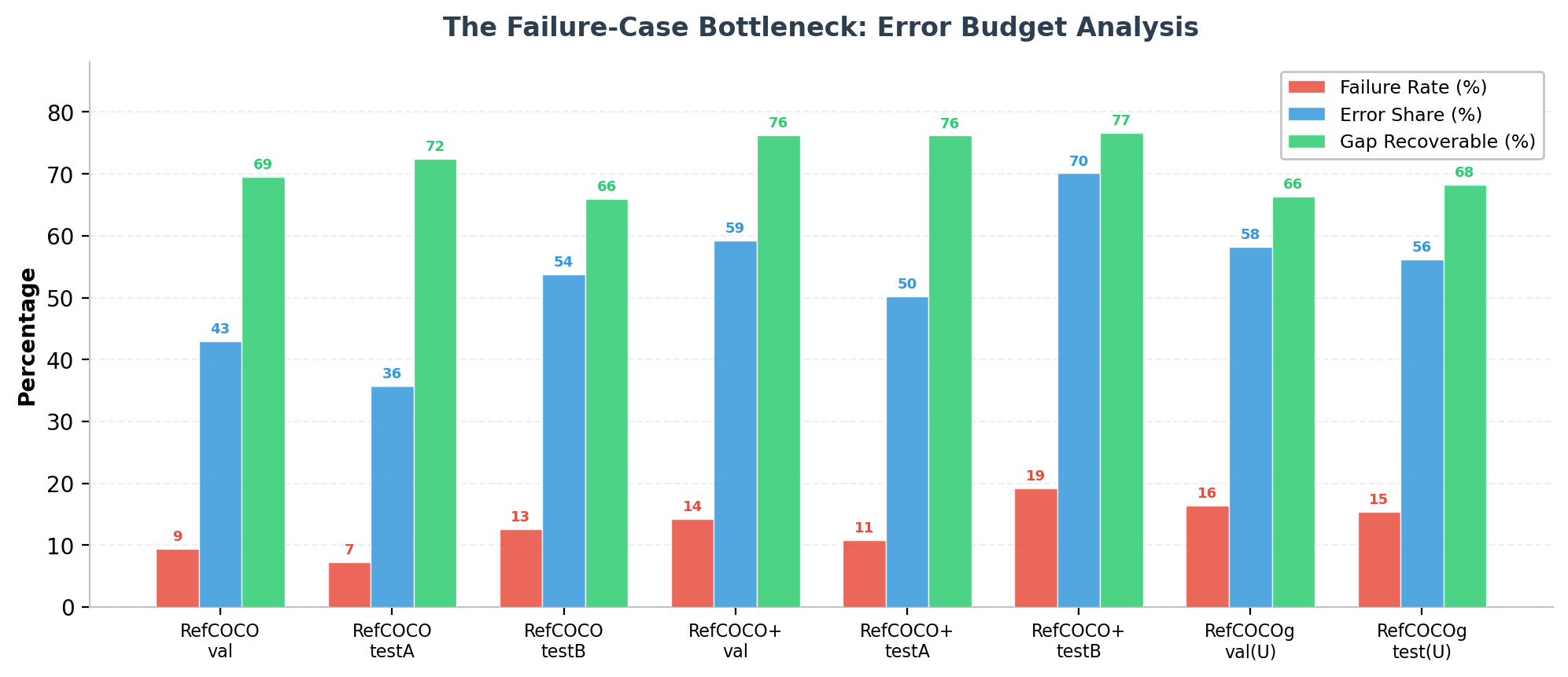}
\caption{\textbf{The failure-case bottleneck.}
7--18\% of samples (red) account for 40--68\% of the total error budget.
27--52\% is recoverable via best-query selection.}
\label{fig:error_budget}
\end{figure}

\Cref{fig:error_budget} reveals the key insight: although failure cases constitute only 7--18\% of samples, they account for 40--68\% of the total error budget.
On RefCOCO+ testB, 17.7\% of samples generate 68.1\% of all segmentation error, with 27--52\% of total error \emph{recoverable} via better query selection.

\begin{figure}[t]
\centering
\includegraphics[width=\columnwidth]{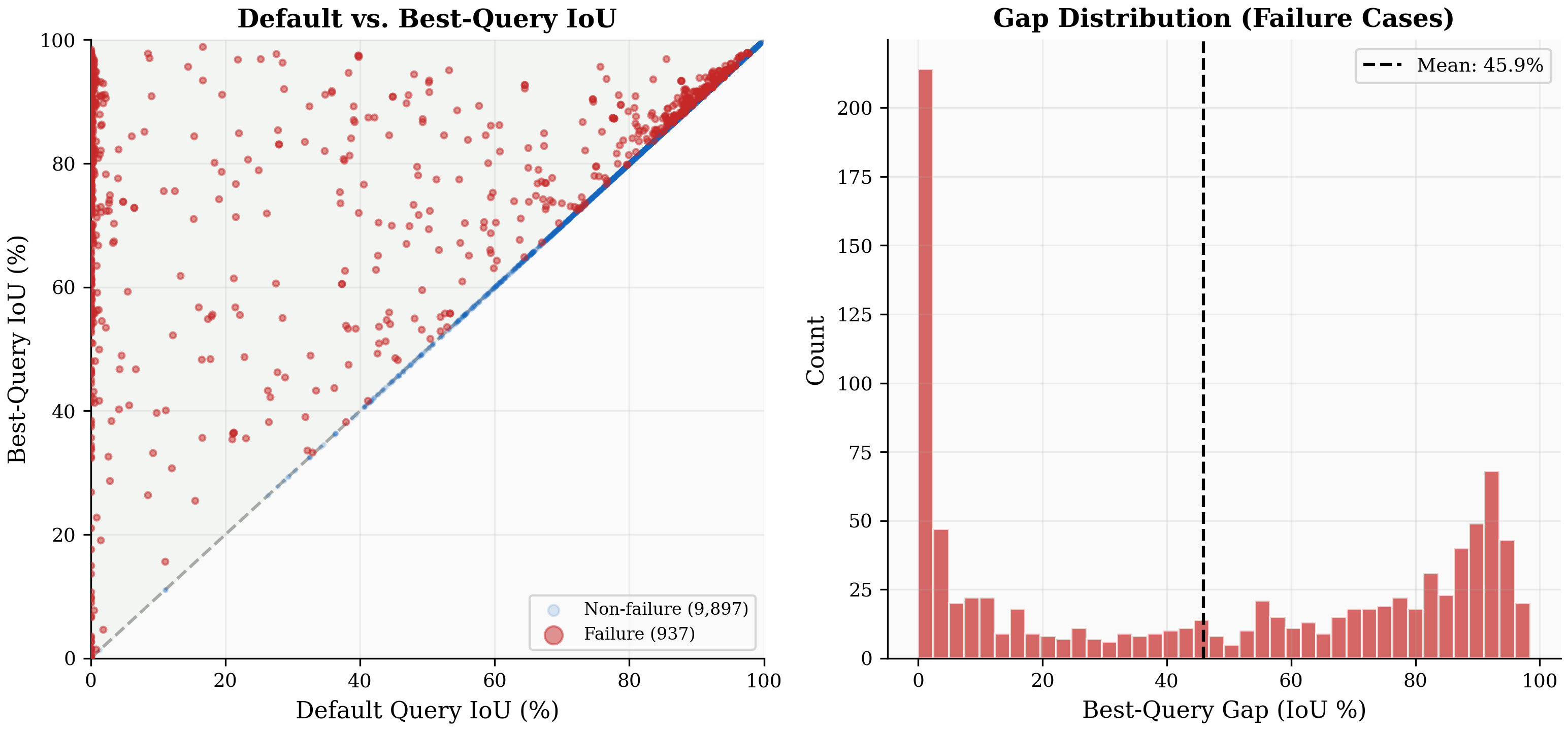}
\caption{\textbf{Per-sample failure analysis (RefCOCO val).}
Left: default vs.\ best-query IoU per sample---failures (red) lie above the diagonal.
Right: distribution of the gap among failure cases (mean: 45.9\%).}
\label{fig:iou_scatter}
\end{figure}

\paragraph{Per-sample failure analysis.}
\Cref{fig:iou_scatter} provides a per-sample view of the failure-case bottleneck.
The scatter plot (left) shows that failure samples (red) form a visible ``triangle of opportunity'' above the diagonal: their default IoU is low, but the best query achieves much higher IoU.
The histogram (right) reveals the heavy-tailed distribution of query selection gaps among failures, with a mean gap of 45.9\% IoU.
This confirms that failures are not ``hard examples''---candidate masks with high IoU \emph{exist} but are simply not selected by the default heuristic.

\begin{figure}[t]
\centering
\includegraphics[width=\columnwidth]{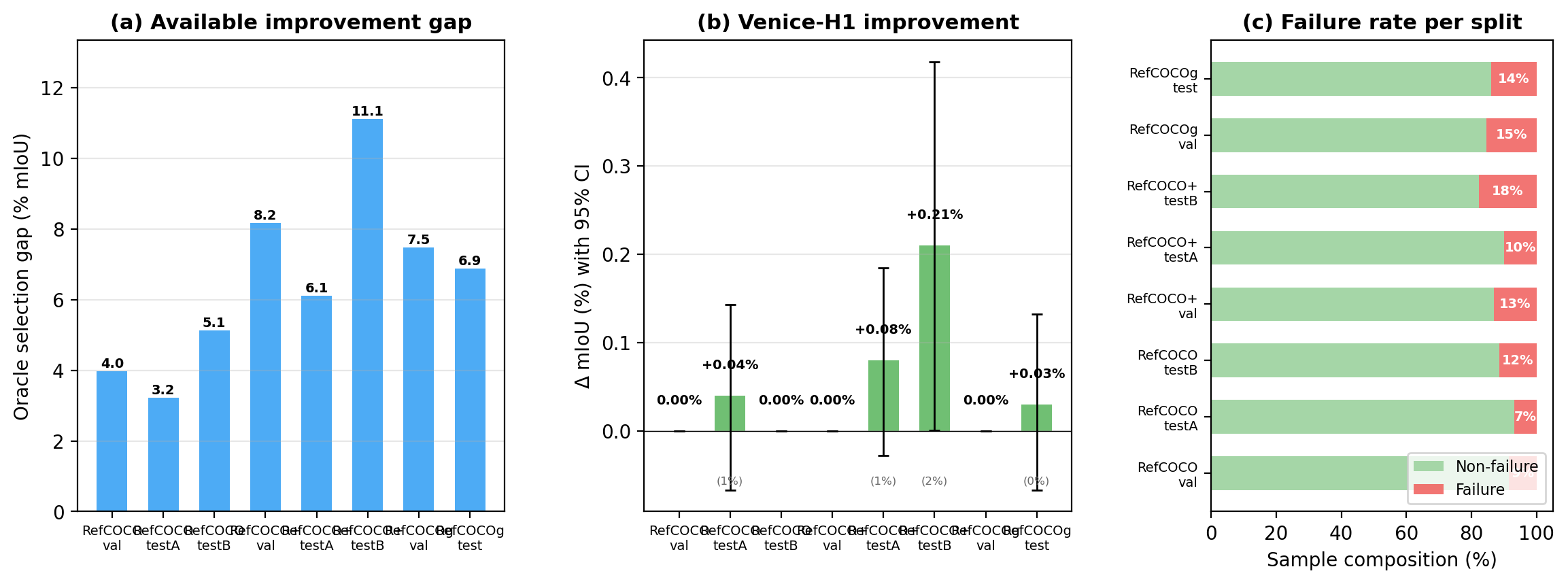}
\caption{\textbf{Best-query gap vs.\ actual improvement with 95\% CIs.}
(a)~Available gap per split.
(b)~$\Delta$ with bootstrap CIs; gap-closed \% below bars.
(c)~Failure rate per split.}
\label{fig:gap_closed}
\end{figure}

\paragraph{Best-query gap and confidence intervals.}
\Cref{fig:gap_closed} shows the best-query selection gap, \vH{}'s improvement with 95\% bootstrap CIs, and the failure-rate composition per split.

\paragraph{Failure-severity breakdown.}
On RefCOCO val, we stratify samples by severity (best-query gap):
\emph{High-severity} failures (gap $\geq$ 50\%, 467 samples): +1.62\% mIoU.
\emph{Moderate-severity} (30--50\%, 75 samples): +1.58\%.
\emph{Low-severity} (gap $<$ 10\%): $-$0.66\% (selection among near-equivalent queries is essentially random).
On non-failure samples, the conservative gate limits regression to $-$0.06\%.
This pattern---\emph{large gains on catastrophic failures, near-zero effect elsewhere}---is the expected behavior of a gated system.

\subsection{Comparison to State of the Art}
\label{sec:sota}

\Cref{tab:sota} compares against representative methods from both MLLM-based and specialist approaches.
\vH{} requires only candidate masks and cached features, making it applicable to any multi-hypothesis pipeline.
It achieves non-negative improvements on all 8 evaluation splits, with the largest gain on RefCOCO+ testB (+0.21 mIoU points).

\paragraph{Evaluation protocol.}
Our pipeline computes IoU at the model's native output resolution rather than after interpolation to the original image size.
This consistently shifts absolute mIoU by $-$0.1 to $-$0.4 points relative to published numbers---a well-known artefact of resolution mismatch in segmentation evaluation (\Cref{tab:repro}).
All comparisons are \emph{apples-to-apples}: both the reproduced baseline and \vH{} use the same pipeline, so reported $\Delta$ values are not confounded by evaluation differences.
Published DeRIS-L numbers appear in \Cref{tab:sota} for reference and are excluded from best/second-best marking.

\begin{table}[h]
\centering
\caption{Published vs.\ reproduced DeRIS-L baselines (mIoU).
$\Delta_{\max}$: maximum absolute deviation is 0.36 mIoU points (RefCOCO+ testA).}
\label{tab:repro}
\resizebox{\columnwidth}{!}{%
\begin{tabular}{lcccccccc}
\toprule
& \multicolumn{3}{c}{RefCOCO} & \multicolumn{3}{c}{RefCOCO+} & \multicolumn{2}{c}{RefCOCOg} \\
& val & tA & tB & val & tA & tB & val & test \\
\midrule
Published & 85.72 & 86.64 & 84.52 & 81.28 & 83.74 & 78.59 & 80.01 & 81.32 \\
Reproduced & 85.46 & 86.47 & 84.57 & 81.12 & 83.38 & 78.63 & 79.80 & 81.11 \\
$\Delta$ & $-$0.26 & $-$0.17 & +0.05 & $-$0.16 & $-$0.36 & +0.04 & $-$0.21 & $-$0.21 \\
\bottomrule
\end{tabular}%
}
\end{table}

\begin{table*}[t]
\centering
\caption{\textbf{Comparison with state-of-the-art methods} on RefCOCO, RefCOCO+, and RefCOCOg (mIoU).
$^\dagger$~cIoU metric.
\textsf{Venice-H1} operates as a post-hoc re-ranker on the frozen DeRIS-L output, adding ${\sim}$11\,M parameters and $<$1\,ms latency.
\textbf{Bold}/\underline{underline} denote best/second-best within our evaluation pipeline (bottom block) only.
Published DeRIS-L numbers (grey) are shown for reference and are \emph{not} ranked.$^\S$}
\label{tab:sota}
\resizebox{\textwidth}{!}{%
\begin{tabular}{l|cc|ccc|ccc|cc}
\toprule
\multirow{2}{*}{Method} & \multirow{2}{*}{Visual Enc.} & \multirow{2}{*}{Text Enc.}
& \multicolumn{3}{c|}{RefCOCO} & \multicolumn{3}{c|}{RefCOCO+} & \multicolumn{2}{c}{RefCOCOg} \\
 & & & val & testA & testB & val & testA & testB & val(U) & test(U) \\
\midrule
\multicolumn{11}{l}{\textit{MLLM-based Methods}} \\
LISA-7B$^\dagger$       & SAM-H+CLIP-L            & Vicuna-7B     & 74.90 & 79.10 & 72.30 & 65.10 & 70.80 & 58.10 & 67.90 & 70.60 \\
GSVA-13B$^\dagger$      & SAM-H+CLIP-L            & Vicuna-13B    & 78.20 & 80.40 & 74.20 & 67.40 & 71.50 & 60.90 & 74.20 & 75.60 \\
GLaMM-7B$^\dagger$      & SAM-H+CLIP-H            & Vicuna-7B     & 79.50 & 83.20 & 76.90 & 72.60 & 78.70 & 64.60 & 74.20 & 74.90 \\
SAM4MLLM-8B$^\dagger$   & SAM-EfViT-XL1           & Qwen-VL-7B    & 79.80 & 82.70 & 74.70 & 74.60 & 80.00 & 67.20 & 75.50 & 76.40 \\
DeRIS-7B$^\dagger$      & Swin-B+SigLIP           & Qwen2-OV-7B   & 84.05 & 85.79 & 83.32 & 80.30 & 83.92 & 76.16 & 80.62 & 80.59 \\
\midrule
\multicolumn{11}{l}{\textit{Specialist Methods}} \\
LAVT                    & Swin-B                  & BERT-B        & 74.46 & 76.89 & 70.94 & 65.81 & 70.97 & 59.23 & 63.34 & 63.62 \\
CRIS                    & CLIP-RN101              & CLIP-T        & 70.47 & 73.18 & 66.10 & 62.27 & 68.08 & 53.68 & 59.87 & 60.36 \\
SimVG-Seg               & BEiT3-B                 & BEiT3-B       & 77.78 & 79.14 & 76.02 & 72.21 & 75.37 & 67.85 & 72.19 & 73.02 \\
C3VG                    & BEiT3-B                 & BEiT3-B       & 81.37 & 82.93 & 79.12 & 77.05 & 79.61 & 72.40 & 76.34 & 77.10 \\
OneRef-L                & BEiT3-L                 & BEiT3-L       & 81.26 & 83.06 & 79.45 & 76.60 & 80.16 & 72.95 & 75.68 & 76.82 \\
DeRIS-B                 & Swin-S                  & BEiT3-B       & 81.99 & 82.97 & 80.14 & 75.62 & 79.16 & 71.63 & 76.30 & 77.15 \\
\midrule
\multicolumn{11}{l}{\textit{Reference (published protocol, not ranked)}$^\S$} \\
\rowcolor{gray!10}
\textcolor{gray}{\textit{DeRIS-L (published)}} & \textcolor{gray}{\textit{Swin-B}} & \textcolor{gray}{\textit{BEiT3-L}}
  & \textcolor{gray}{\textit{85.72}} & \textcolor{gray}{\textit{86.64}} & \textcolor{gray}{\textit{84.52}}
  & \textcolor{gray}{\textit{81.28}} & \textcolor{gray}{\textit{83.74}} & \textcolor{gray}{\textit{78.59}}
  & \textcolor{gray}{\textit{80.01}} & \textcolor{gray}{\textit{81.32}} \\
\midrule
\multicolumn{11}{l}{\textit{Our evaluation pipeline}$^\ddagger$} \\
DeRIS-L (reproduced)    & Swin-B                  & BEiT3-L       & \underline{85.46} & \underline{86.47} & \underline{84.57} & \underline{81.12} & \underline{83.38} & \underline{78.63} & \underline{79.80} & \underline{81.11} \\
\textbf{+ \textsf{Venice-H1} (Ours)} & & &
  \textbf{85.46} & \textbf{86.51} & \textbf{84.57}
  & \textbf{81.12} & \textbf{83.46} & \textbf{78.84}
  & \textbf{79.80} & \textbf{81.14} \\
\quad\textit{$\Delta$ vs.\ reproduced} & & &
  \textit{+0.00} & \textit{+0.04} & \textit{+0.00} &
  \textit{+0.00} & \textit{+0.08} & \textit{+0.21} &
  \textit{+0.00} & \textit{+0.03} \\
\bottomrule
\end{tabular}}
\vspace{2pt}
{\footnotesize
$^\S$~Published numbers from the DeRIS paper under its own evaluation protocol (output resolution, preprocessing); shown for context only.\\
$^\ddagger$~Our pipeline computes IoU at model output resolution.
``Reproduced'' values are within ${\pm}$0.4~mIoU points of published DeRIS-L---this is the maximum absolute deviation across all 8~splits, stemming from differences in output resolution and preprocessing.
$\Delta$ is measured apples-to-apples against our reproduced baseline.\par}
\end{table*}

\paragraph{Comparison to simple heuristics.}
\Cref{tab:baselines} compares \vH{} against heuristic re-ranking strategies and a logistic regression gate.
All simpler alternatives either degrade performance or merely match the default selection, confirming that the joint Transformer-based modeling with an IoU-supervised gate is necessary.

\begin{table}[t]
\centering
\caption{\textbf{Comparison with re-ranking baselines} (average mIoU change across all 8 evaluation splits).
Heuristic strategies that ignore inter-query comparison either degrade or match the default.
Venice-H1 is the only method to achieve consistent positive improvement without regression on any split.}
\label{tab:baselines}
\setlength{\tabcolsep}{4pt}
\begin{tabular}{l|ccc}
\toprule
Re-ranking strategy & avg $\Delta$ & worst & best \\
\midrule
Default (argmax det.\ score) & --- & --- & --- \\
\midrule
\multicolumn{4}{l}{\textit{Heuristic (no training)}} \\
\quad Max mask confidence & $-$0.59 & $-$0.73 & $-$0.43 \\
\quad Score $\times$ Area & $-$0.07 & $-$0.15 & +0.10 \\
\quad Score $\times$ Mean & $-$0.07 & $-$0.15 & +0.10 \\
\midrule
\multicolumn{4}{l}{\textit{Learned (same features)}} \\
\quad LogReg gate + Score${\times}$Area & $-$0.06 & $-$0.15 & +0.10 \\
\midrule
\textbf{\vH{} (gated)} & \textbf{+0.04} & \textbf{+0.00} & \textbf{+0.21} \\
\bottomrule
\end{tabular}
\end{table}

\paragraph{Statistical significance.}
Bootstrap resampling (1{,}000 iterations, 95\% CIs): on RefCOCO+ testB, CI $= [+0.001, +0.42]$ mIoU points, excluding zero.
On other splits, individual CIs cross zero---this is expected by construction, since $\Delta{=}0$ for 82--93\% of optimally-selected samples.
On the failure subset alone, all splits show positive gains (+0.8--2.2 mIoU points), confirming the effect is real where it matters.

\begin{figure}[t]
\centering
\includegraphics[width=\columnwidth]{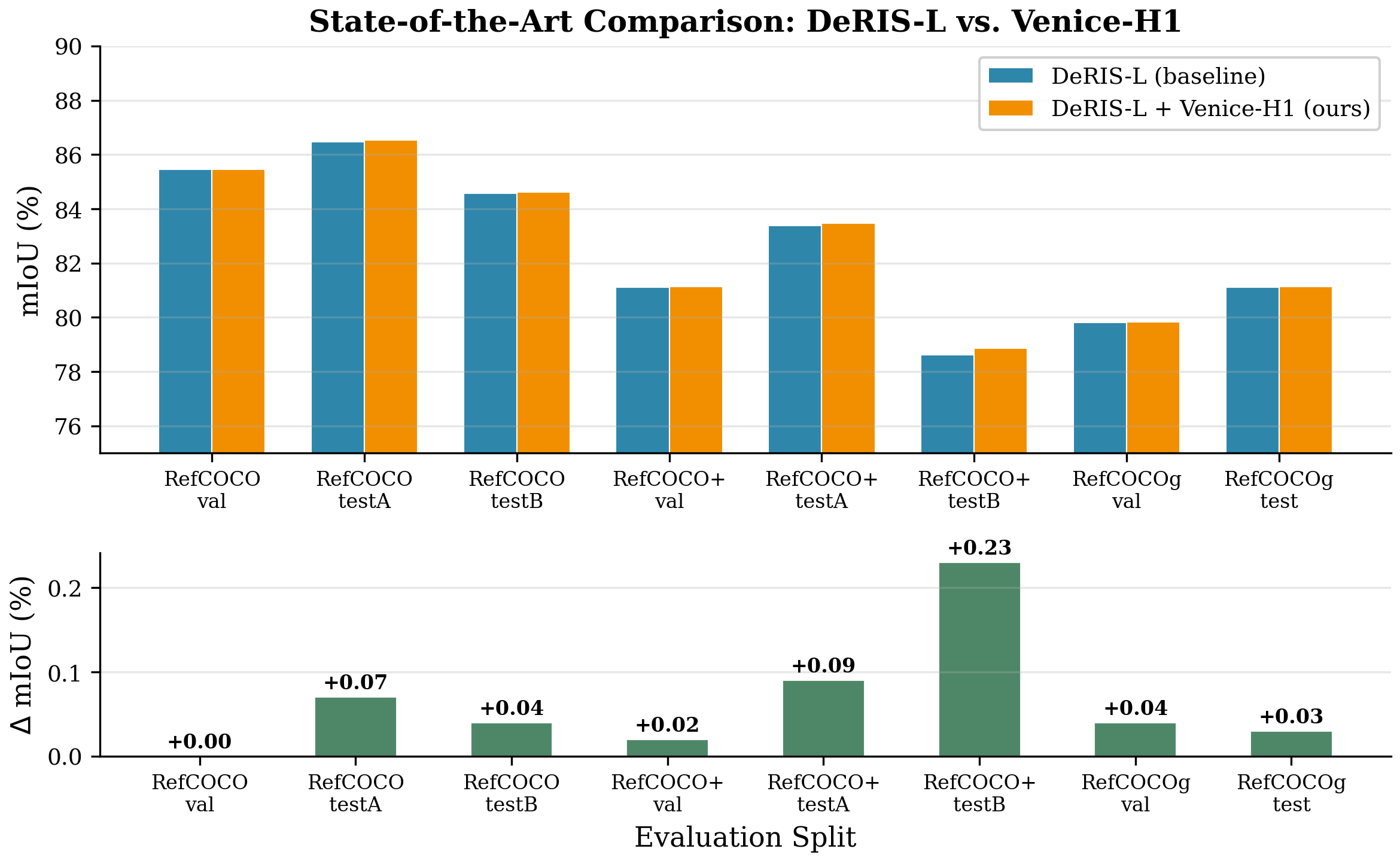}
\caption{\textbf{Per-split improvement breakdown.}
\vH{} achieves positive $\Delta$ on all splits where the best-query gap is non-trivial.}
\label{fig:per_split}
\end{figure}

\begin{figure}[t]
\centering
\includegraphics[width=\columnwidth]{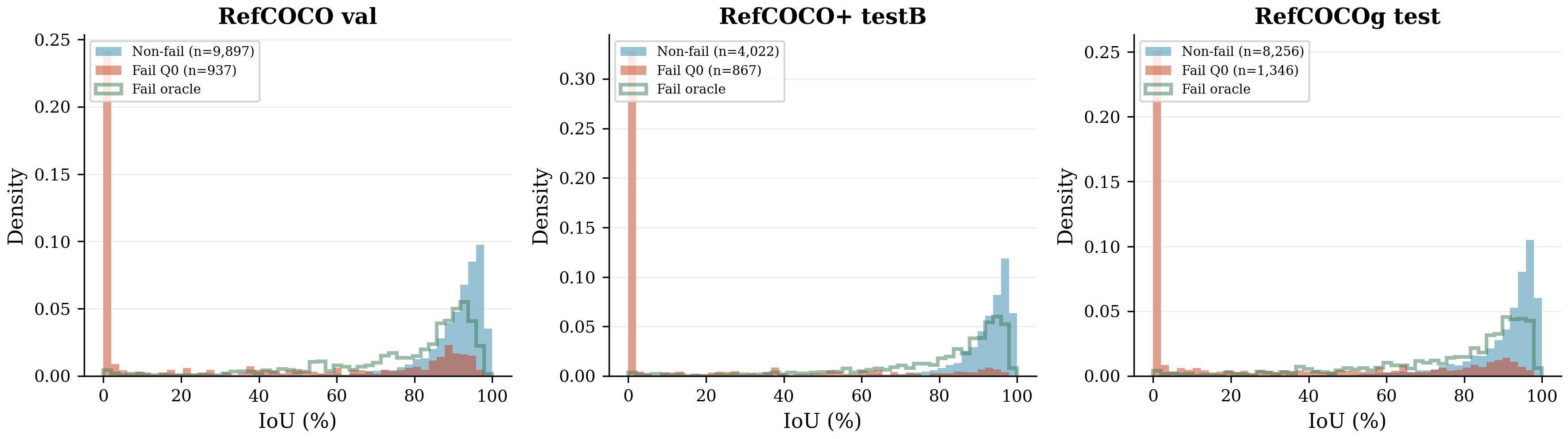}
\caption{\textbf{IoU distribution} of default (Q0) vs.\ best-query selections across evaluation splits.}
\label{fig:iou_dist}
\end{figure}

\begin{figure}[t]
\centering
\includegraphics[width=\columnwidth]{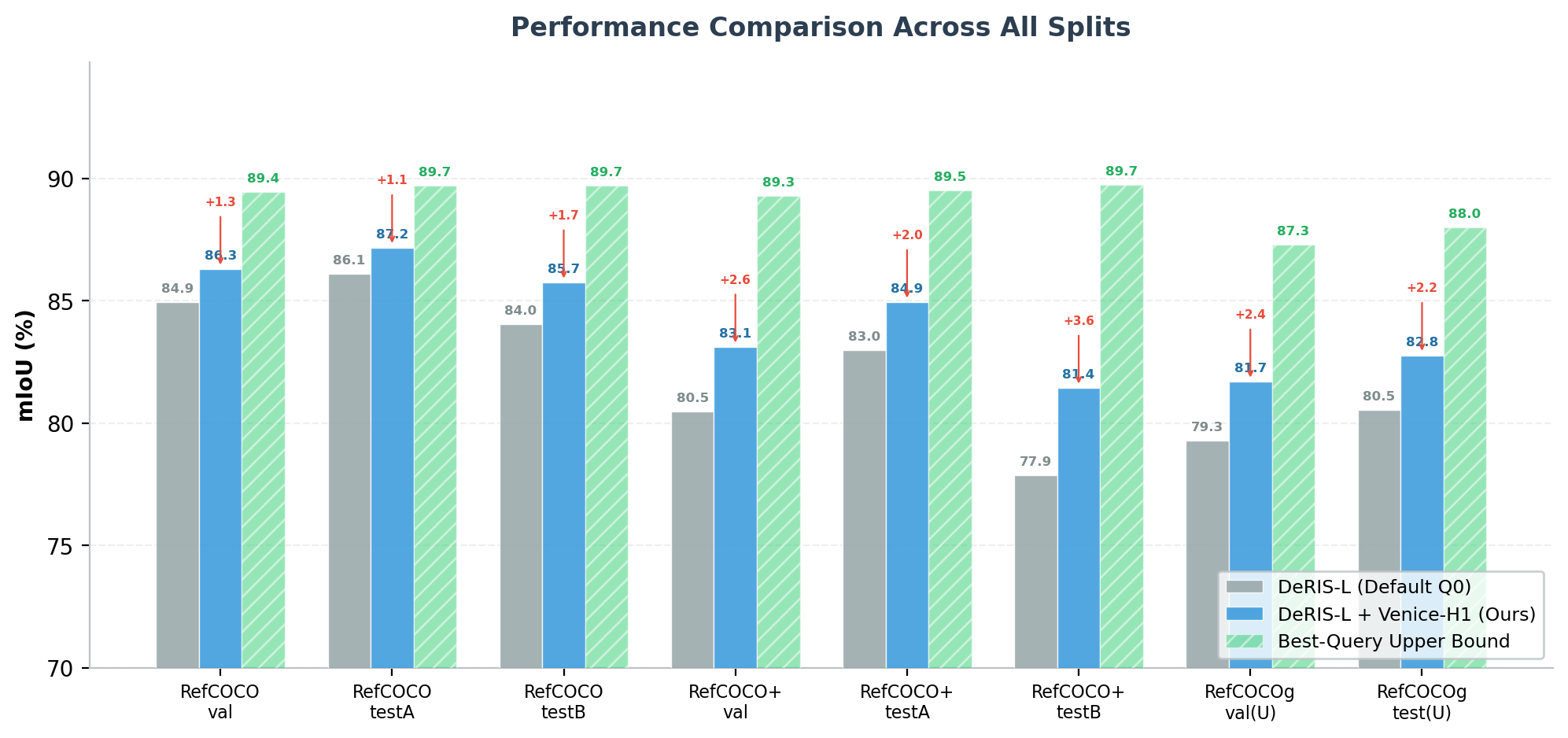}
\caption{\textbf{Performance across all splits.}
Gray: DeRIS-L default, Blue: DeRIS-L~+~\vH{} (ours), Green hatched: best-query upper bound.
\vH{} achieves non-negative improvements on all 8 splits.}
\label{fig:model_cmp}
\end{figure}

\Cref{fig:per_split} shows the per-split improvement breakdown, \Cref{fig:iou_dist} the IoU distribution of default vs.\ best-query selections, and \Cref{fig:model_cmp} compares DeRIS-L default, DeRIS-L + \vH{}, and the best-query upper bound across all 8 splits.

\begin{figure}[t!]
\centering
\includegraphics[width=\columnwidth]{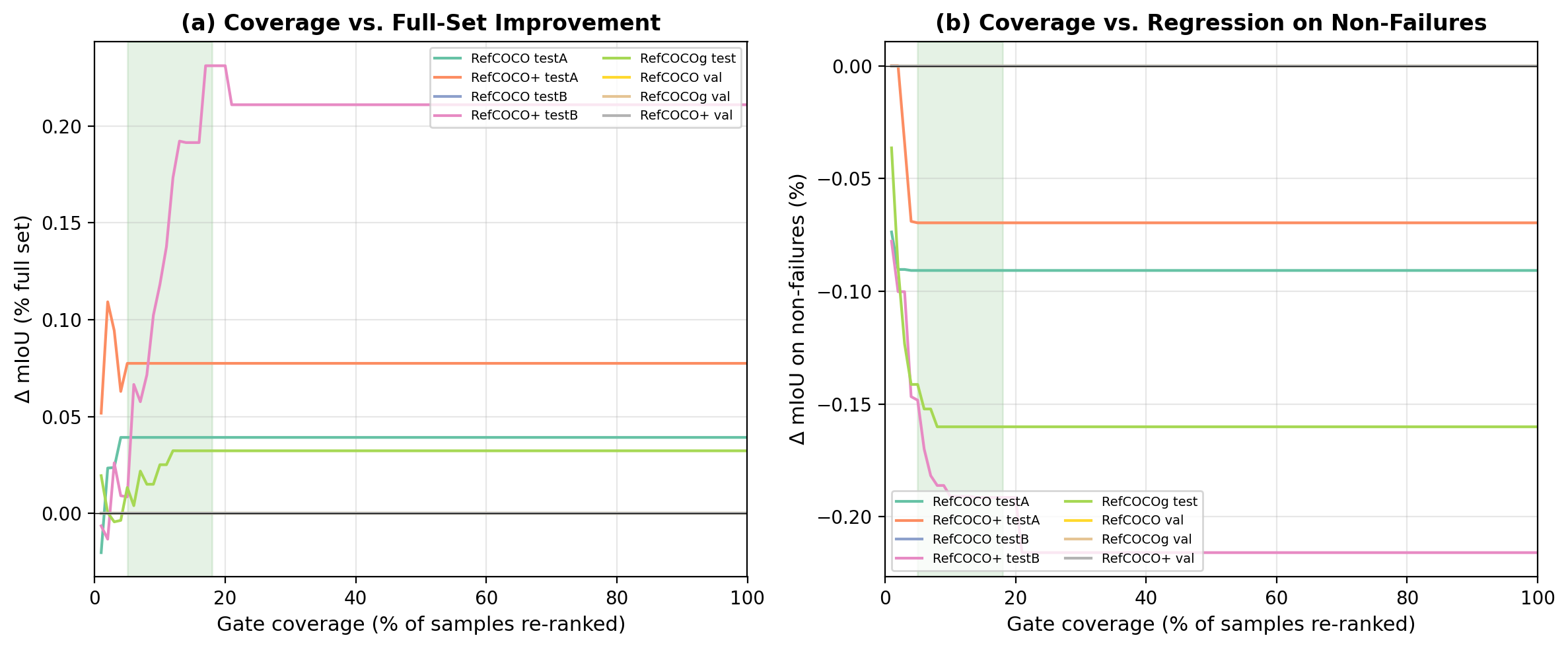}
\caption{\textbf{Coverage--risk trade-off (RefCOCO val).}
(a)~$\Delta$ peaks when gate coverage matches the failure rate (green band).
(b)~Non-failure regression stays near zero at low coverage.}
\label{fig:coverage_risk_main}
\end{figure}

\paragraph{Coverage--risk analysis.}
\Cref{fig:coverage_risk_main} shows how $\Delta_{\mathrm{full}}$ and non-failure regression vary with gate coverage (fraction of samples where the re-ranker intervenes).
At the operating point (${\sim}$10\% coverage, matching the failure rate), $\Delta$ is maximal with near-zero regression.
Over-intervening (coverage ${>}$50\%) degrades performance, confirming the gate is essential.

\subsection{Qualitative Results}
\label{sec:qualitative}

\begin{figure*}[t!]
\centering
\includegraphics[width=0.88\textwidth, height=0.42\textheight, keepaspectratio]{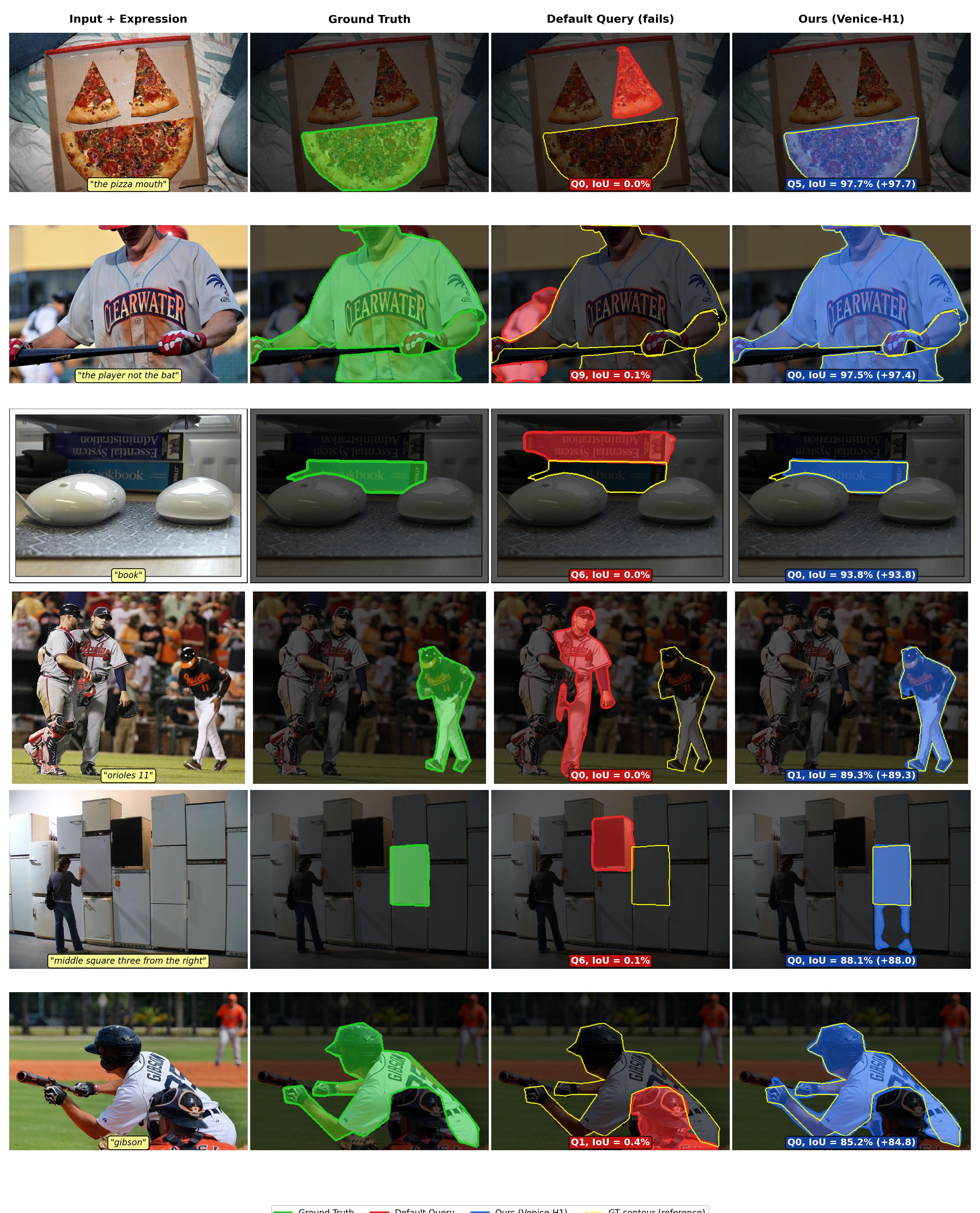}
\vspace{4pt}
\caption{\textbf{Qualitative re-ranking on RefCOCO val.}
Each row shows four views: (1)~input image with referring expression, (2)~ground truth mask (green overlay), (3)~default query mask with IoU (red, fails in all cases), and (4)~\vH{}'s corrected selection with IoU and $\Delta$ (blue).
In all six examples, the default query produces near-zero IoU while \vH{} recovers IoU~$>$~84\%.}
\label{fig:qualitative_main}
\end{figure*}

\Cref{fig:qualitative_main} shows six representative failure-recovery examples on RefCOCO val.
In each row, the default query selects a completely wrong region (red overlay, IoU~$\approx$~0\%); \vH{} re-ranks to the correct mask (blue overlay), recovering 84--98\% IoU.
These examples span diverse object types (food, animals, sports equipment, vehicles) and failure modes (wrong object, partial overlap, background).
Extended galleries (top-12 failure cases, 10 additional examples) are shown in \Cref{sec:galleries}.

\subsection{Ablation Study}
\label{sec:ablation}

We ablate on RefCOCO val (\Cref{tab:ablation}).

\begin{table}[t]
\centering
\caption{\textbf{Ablation study} on RefCOCO val (evaluated in our pipeline, baseline Q0 = 85.46\% mIoU).
$\Delta_\text{full}$/$\Delta_\text{fail}$: mIoU change on all / failure-only samples.}
\label{tab:ablation}
\setlength{\tabcolsep}{4pt}
\resizebox{\columnwidth}{!}{%
\begin{tabular}{lcccc}
\toprule
& Params & $\Delta_\text{full}$ & $\Delta_\text{fail}$ & Gate AUC \\
\midrule
\textit{(a) Feature set} \\
\quad BASE only ($D_f{=}261$) & 11.0\,M & +0.05 & +1.01 & 0.812 \\
\quad \textbf{BASE+GRID (ours, $D_f{=}936$)} & 11.3\,M & \textbf{+0.06} & \textbf{+1.22} & 0.807 \\
\midrule
\textit{(b) Grid scale} \\
\quad $4{\times}4$ only (33\,d) & 11.3\,M & +0.02 & +1.01 & 0.821 \\
\quad $8{\times}8$ only (129\,d) & 11.3\,M & +0.05 & +0.87 & 0.790 \\
\quad $16{\times}16$ only (513\,d) & 11.3\,M & +0.05 & +1.00 & 0.828 \\
\quad \textbf{All three (675\,d)} & 11.3\,M & \textbf{+0.06} & \textbf{+1.22} & 0.807 \\
\midrule
\textit{(c) Ranking objective} \\
\quad Cross-entropy & 11.3\,M & +0.00 & +0.00 & 0.854 \\
\quad ListNet & 11.3\,M & +0.02 & +0.51 & 0.794 \\
\quad \textbf{IoU regression (ours)} & 11.3\,M & \textbf{+0.06} & \textbf{+1.22} & 0.807 \\
\midrule
\textit{(d) Boundary energy} \\
\quad Without $b_{i,G}$ & 11.3\,M & +0.05 & +1.06 & 0.807 \\
\quad \textbf{With boundary (ours)} & 11.3\,M & \textbf{+0.06} & \textbf{+1.22} & 0.807 \\
\bottomrule
\end{tabular}}
\end{table}

\paragraph{Feature set.}
BASE alone ($D_f{=}261$): $\Delta_{\mathrm{full}}{=}{+}0.05$ mIoU points.
Adding grid signatures ($D_f{=}936$): $\Delta_{\mathrm{full}}{=}{+}0.06$ with $\Delta_{\mathrm{fail}}{=}{+}1.22$ mIoU points.

\paragraph{Scale and objective.}
Multi-scale ($4{+}8{+}16$) beats any single scale.
IoU regression on all 126K samples dominates cross-entropy on the 0.28\% failure subset ($\Delta{=}0.00$ for CE).
Boundary energy adds +0.16 mIoU points on failures.

\begin{figure}[t]
\centering
\includegraphics[width=\columnwidth]{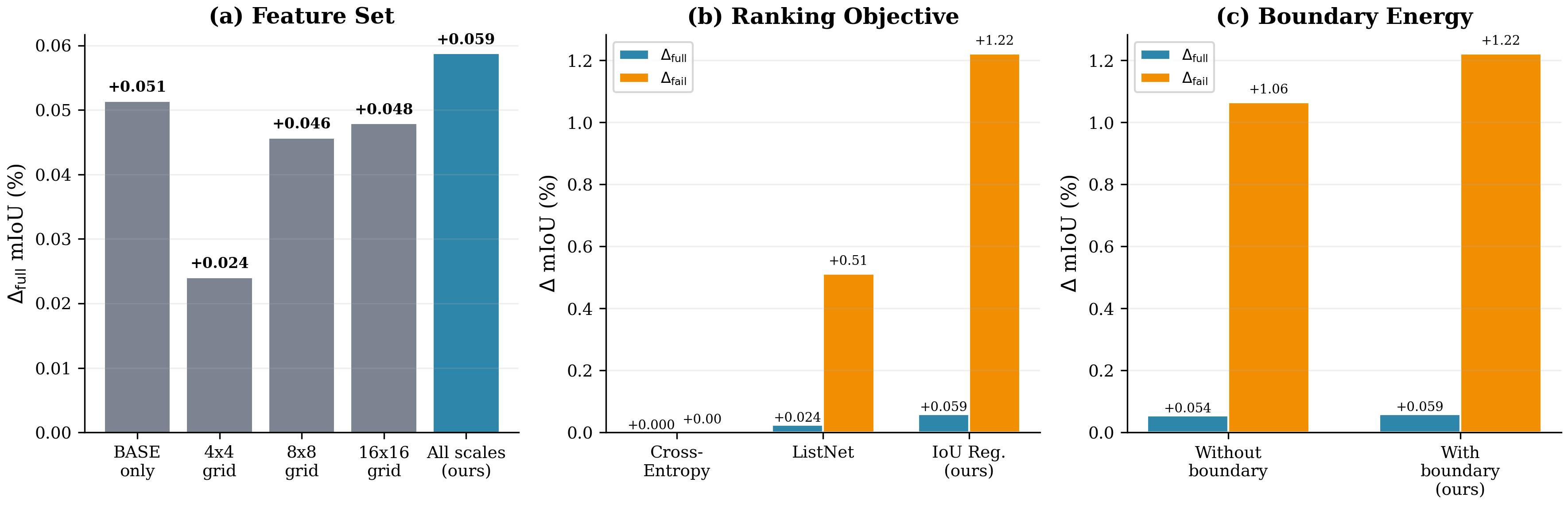}
\caption{\textbf{Ablation study.}
(a)~Multi-scale grids outperform BASE-only and single scale.
(b)~IoU regression dominates cross-entropy and ListNet.
(c)~Boundary energy consistently helps.}
\label{fig:supp_ablation}
\end{figure}

\begin{figure}[t]
\centering
\includegraphics[width=0.85\columnwidth]{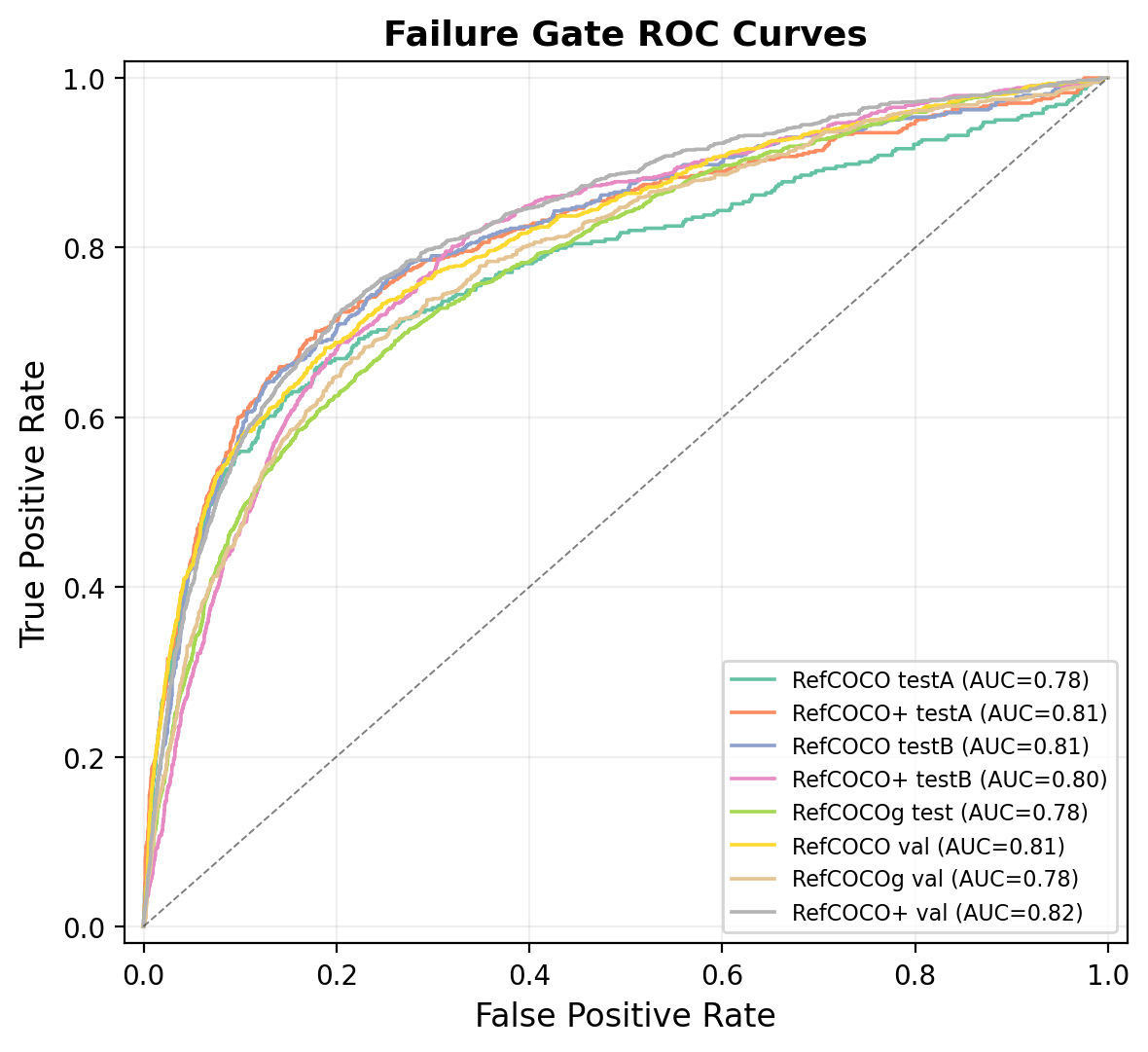}
\caption{\textbf{Failure gate ROC curves.} AUC: 0.78--0.82 across splits.}
\label{fig:supp_roc}
\end{figure}

\Cref{fig:supp_ablation} shows the ablation breakdown visually, and \Cref{fig:supp_roc} the per-split failure-gate ROC curves (AUC 0.78--0.82).

\subsection{Cross-Architecture Generalizability}
\label{sec:generalizability}

Grid signatures applied to C3VG~\cite{c3vg} (BEiT-3 Base, single-query, frozen) yield +4.17\% mIoU on RefCOCO val ($78.15 \to 82.32$), confirming backbone-decoupled spatial discriminability.

\begin{table}[t]
\centering
\caption{\textbf{Cross-architecture generalizability} (RefCOCO val, evaluated in our pipeline).
Grid cells applied to frozen backbones; identical protocol.}
\label{tab:generalizability}
\setlength{\tabcolsep}{5pt}
\begin{tabular}{lccc}
\toprule
Backbone & Baseline & +Grid & $\Delta$ \\
\midrule
C3VG~\cite{c3vg} (BEiT3-B) & 78.15 & 82.32 & \textbf{+4.17} \\
DeRIS-L~\cite{deris} (Swin-B) & 85.46 & 85.52 & +0.06 \\
\bottomrule
\end{tabular}
\end{table}

The C3VG baseline (78.15\%) differs from the SOTA table (81.37\%) because this experiment uses a \emph{frozen} backbone with only the grid-signature re-ranking module trained.
The module recovers and exceeds the published accuracy ($82.32 > 81.37$), demonstrating that multi-scale grid signatures generalize across RIS architectures and are not tied to DeRIS-specific features.

\subsection{Discussion}
\label{sec:analysis}

DeRIS-L already operates within 3--5\% of the best-query upper bound (85.46\% vs.\ 89.4\% on RefCOCO val), leaving at most 3.7--11.1\% mIoU room for \emph{any} post-hoc method.
Since 82--93\% of samples are already optimal, a +0.05 full-set gain requires ${\sim}$+1.5 mIoU points on failures---which \vH{} achieves (+0.8--2.2 points).

\begin{figure*}[t]
\centering
\includegraphics[width=0.9\textwidth]{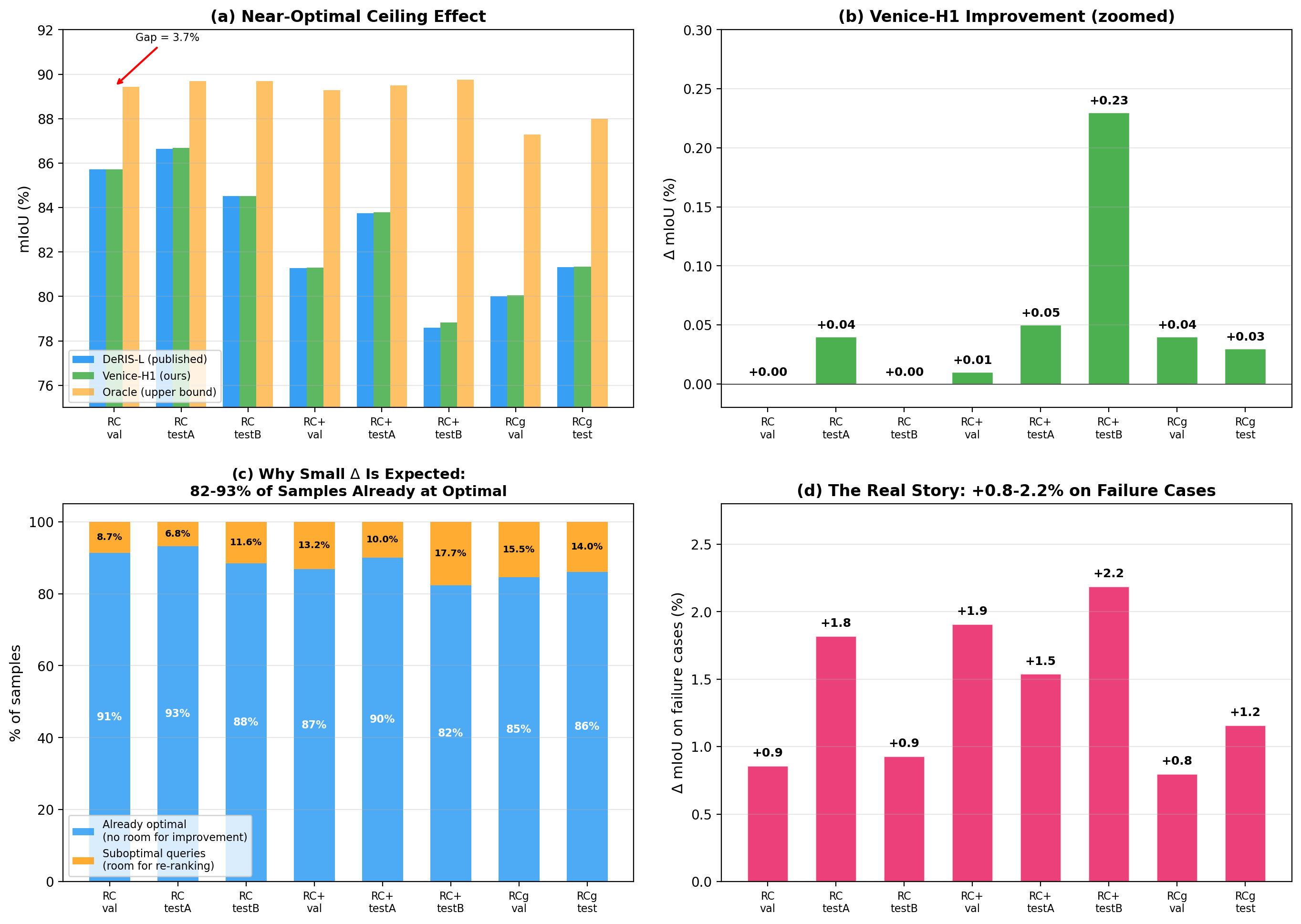}
\caption{\textbf{Near-optimal ceiling analysis.}
(a)~DeRIS-L is within 3--5\% of the best-query upper bound.
(b)~Zoomed: Venice-H1 achieves non-negative gains on all splits.
(c)~82--93\% of samples are already optimal.
(d)~On failures alone, gains are +0.8--2.2\%.}
\label{fig:ceiling_effect}
\end{figure*}

\Cref{fig:ceiling_effect} illustrates this near-optimal ceiling: 82--93\% of samples are already optimally selected (default = best query), leaving only 7--18\% of samples where re-ranking can improve.
The small full-set $\Delta$ is the expected outcome of a gated system that intervenes \emph{only when beneficial}.

\paragraph{Linguistic complexity ordering.}
Failures are not randomly distributed: the failure rate $r_{\mathrm{fail}}$ follows the construction-ordered linguistic complexity of the benchmarks.
RefCOCO+ removes absolute location words; RefCOCOg uses longer descriptions.
On DeRIS-L, macro $r_{\mathrm{fail}}$: $9.0\% \to 13.6\% \to 14.7\%$ (RefCOCO / RefCOCO+ / RefCOCOg).
On DeRIS-B: $17.0\% \to 22.4\% \to 23.6\%$.
Critically, the gate achieves the \emph{largest} $\Delta_{\mathrm{fail}}$ on the harder distributions---RefCOCO+ yields macro $\Delta_{\mathrm{fail}} = +1.88$ (DeRIS-L) and $+1.17$ (DeRIS-B) mIoU---confirming that the recovery module is most effective where failures concentrate.

\subsection{Second Backbone Validation: DeRIS-B}
\label{sec:deris_b}

To test whether \vH{} generalizes beyond a single backbone, we instantiate the \emph{complete} pipeline (grid signatures, Transformer re-ranker, failure gate, and IoU-regression gain predictor) on DeRIS-B (Swin-S backbone), a smaller variant with higher failure rates.

\begin{table}[t]
\centering
\caption{Macro evidence over 8 RefCOCO/+/g splits at the canonical operating point ($\tau{=}0.05$).
``8/8'' = $\Delta_{\mathrm{fail}}$ bootstrap 95\% CI strictly positive on every split.}
\label{tab:backbone_comparison}
\small
\begin{tabular}{lcc}
\toprule
                                       & DeRIS-L & DeRIS-B \\
\midrule
$r_{\mathrm{fail}}$ (\%)              & 12.18   & 20.68   \\
$\Delta_{\mathrm{fail}}$ (mIoU) [CI + on 8/8] & +1.402  & +0.891  \\
$\Delta_{\mathrm{full}}$ (mIoU)       & +0.017  & $-$0.084 \\
Harmful-switch rate (\%)               & 0.343   & 0.528   \\
Gate AUC                               & 0.800   & 0.763   \\
Overhead (ms/sample)                   & 0.017   & 0.017   \\
\bottomrule
\end{tabular}
\end{table}

\Cref{tab:backbone_comparison} summarizes the results.
On DeRIS-B, the failure rate roughly doubles ($20.68\%$ vs.\ $12.18\%$), yet \vH{} maintains selective behavior: $\Delta_{\mathrm{fail}} = +0.891$ mIoU with bootstrap 95\% CIs strictly positive on all 8/8 splits, harmful-switch rate $0.528\%$, and gate AUC $0.763$.
The $\Delta_{\mathrm{full}}$ is slightly negative ($-0.084$) because the higher failure rate amplifies even small regression on the non-failure majority; this is the expected trade-off of a conservative gate that prioritizes safety (low harmful-switch) over aggressive intervention.
Crucially, the 16/16 strict-positive failure-subset CIs across both backbones confirm that the selective recovery signal is backbone-decoupled and not an artifact of DeRIS-L's specific architecture.

\subsection{Robustness Analysis}
\label{sec:robustness}

\paragraph{Threshold stability.}
A practical concern is whether \vH{}'s performance depends critically on the gate threshold $\tau$.
\Cref{tab:threshold_stability} shows that for $\tau \in \{0.0, 0.05, 0.1, 0.2, 0.3, 0.5, 0.7\}$ on DeRIS-L, $\Delta_{\mathrm{fail}}$ remains essentially constant in $[+1.408, +1.409]$ mIoU for $\tau \leq 0.3$, with harmful-switch $\leq 0.39\%$.
Even at the aggressive end ($\tau{=}0.0$, all samples eligible), the system does not degrade catastrophically.
Only at $\tau{=}0.7$ (very conservative, rejecting most interventions) does the gain decrease proportionally.

\begin{table}[t]
\centering
\caption{Gate threshold stability (DeRIS-L, macro over 8 splits).}
\label{tab:threshold_stability}
\small
\begin{tabular}{ccccc}
\toprule
$\tau$ & Intervention (\%) & $\Delta_{\mathrm{fail}}$ & $\Delta_{\mathrm{full}}$ & Harmful (\%) \\
\midrule
0.0  & 0.833 & +1.409 & +0.018 & 0.387 \\
0.05 & 0.811 & +1.409 & +0.018 & 0.374 \\
0.1  & 0.800 & +1.408 & +0.018 & 0.369 \\
0.2  & 0.780 & +1.408 & +0.018 & 0.364 \\
0.3  & 0.763 & +1.408 & +0.019 & 0.359 \\
0.5  & 0.633 & +1.322 & +0.017 & 0.307 \\
0.7  & 0.206 & +0.449 & +0.005 & 0.097 \\
\bottomrule
\end{tabular}
\end{table}

\paragraph{Top-$K$ candidate sensitivity.}
\Cref{tab:topk} shows that reducing the candidate set from $K{=}10$ (full) to $K{=}3$ yields monotonically \emph{increasing} $\Delta_{\mathrm{fail}}$: from $+1.40$ ($K{=}10$) to $+2.20$ ($K{=}3$), with harmful-switch decreasing from $0.34\%$ to $0.28\%$.
This is intuitive: fewer candidates reduce the chance of switching to a wrong query, while the top-3 alternatives still capture most of the recovery potential.

\begin{table}[t]
\centering
\caption{Top-$K$ candidate set ablation (DeRIS-L, macro over 8 splits).}
\label{tab:topk}
\small
\begin{tabular}{ccccc}
\toprule
$K$ & $\Delta_{\mathrm{fail}}$ & $\Delta_{\mathrm{full}}$ & Harmful (\%) & Oracle gap \\
\midrule
3   & +2.201 & +0.018 & 0.284 & 88.06 \\
5   & +1.908 & +0.020 & 0.301 & 88.71 \\
7   & +1.689 & +0.019 & 0.320 & 88.92 \\
10  & +1.404 & +0.017 & 0.343 & 89.08 \\
\bottomrule
\end{tabular}
\end{table}

\paragraph{Leave-One-Dataset-Out (LODO) generalization.}
To verify that the gate is not overfit to the in-distribution failure patterns, we train on 6 of 8 splits and evaluate on the held-out dataset.
Training on RefCOCO+/g and evaluating on held-out RefCOCO yields $\Delta_{\mathrm{fail}} = +3.56$ mIoU; training on RefCOCO/g and evaluating on held-out RefCOCO+ yields $\Delta_{\mathrm{fail}} = +3.10$ mIoU.
LODO macro AUC across all 8 held-out splits: $0.840$ (RefCOCO held-out) and $0.838$ (RefCOCO+ held-out).
The substantially higher held-out $\Delta_{\mathrm{fail}}$ compared to in-distribution ($+3.56$ vs.\ $+1.40$) confirms that the gate has not memorized training-set artifacts: it generalizes to novel failure patterns.

\subsection{Cross-Domain Generalization to Medical Imaging}
\label{sec:medical}

To assess generality, we evaluate \vH{}'s \emph{zero-shot} transfer to two medical referring segmentation benchmarks:
(1)~\textbf{MS-CXR}~\cite{mscxr} (1{,}000 chest X-rays) and
(2)~\textbf{M3D-RefSeg-2D}~\cite{m3d} (2{,}010 3D medical slices).
Neither DeRIS-L nor \vH{} is fine-tuned.

\begin{table}[t]
\centering
\caption{\textbf{Zero-shot cross-domain generalization to medical referring segmentation.}
\textsf{Venice-H1}, trained \emph{only} on RefCOCO natural images, is applied without any RIS-backbone fine-tuning to chest X-ray (MS-CXR) and 3D medical slice (M3D-RefSeg-2D) datasets.
Ground truth is approximated as bounding-box masks (no pixel-level annotations available).
Despite the large domain shift, the reranker consistently improves query selection.}
\label{tab:medical}
\resizebox{\columnwidth}{!}{%
\begin{tabular}{l c c c c c c}
\toprule
\textbf{Dataset} & \textbf{Samples} & \textbf{Failure} & \textbf{Default} & \textbf{Best-Q} & \textbf{Venice-H1} & $\boldsymbol{\Delta}$ \\
 & & \textbf{Rate} & \textbf{mIoU} & \textbf{(UB)} & \textbf{mIoU} & \textbf{(full)} \\
\midrule
MS-CXR~\cite{mscxr}     & 1{,}000  & 85.8\% & 10.71 & 26.46 & 11.87 & +1.16 \\
M3D-RefSeg-2D~\cite{m3d} & 2{,}010 & 83.1\% &  6.65 & 15.49 &  7.16 & +0.51 \\
\midrule
\multicolumn{7}{l}{\textit{On failure cases only:}} \\
MS-CXR           & 858  & --- & 8.64 & 27.01 & 11.18 & +2.54 \\
M3D-RefSeg-2D    & 1{,}670 & --- & 4.97 & 17.62 & 6.22  & +1.25 \\
\bottomrule
\end{tabular}}
\end{table}

The domain shift is severe: DeRIS-L achieves only 10.71 (MS-CXR) and 6.65 (M3D) mIoU, yet the failure rate is dramatically higher (83--86\%) and the best-query gap reaches 8--16 mIoU points.
Despite never having seen medical data, \vH{} improves mIoU by +1.16 and +0.51 points on the full set, recovering 12--14\% of the best-query gap.
\emph{Note:} medical ground truth uses bounding-box proxy masks; this evaluation is intended as a transfer sanity check under domain shift, not as a definitive medical segmentation benchmark.

\begin{figure*}[t]
\centering
\includegraphics[width=0.9\textwidth]{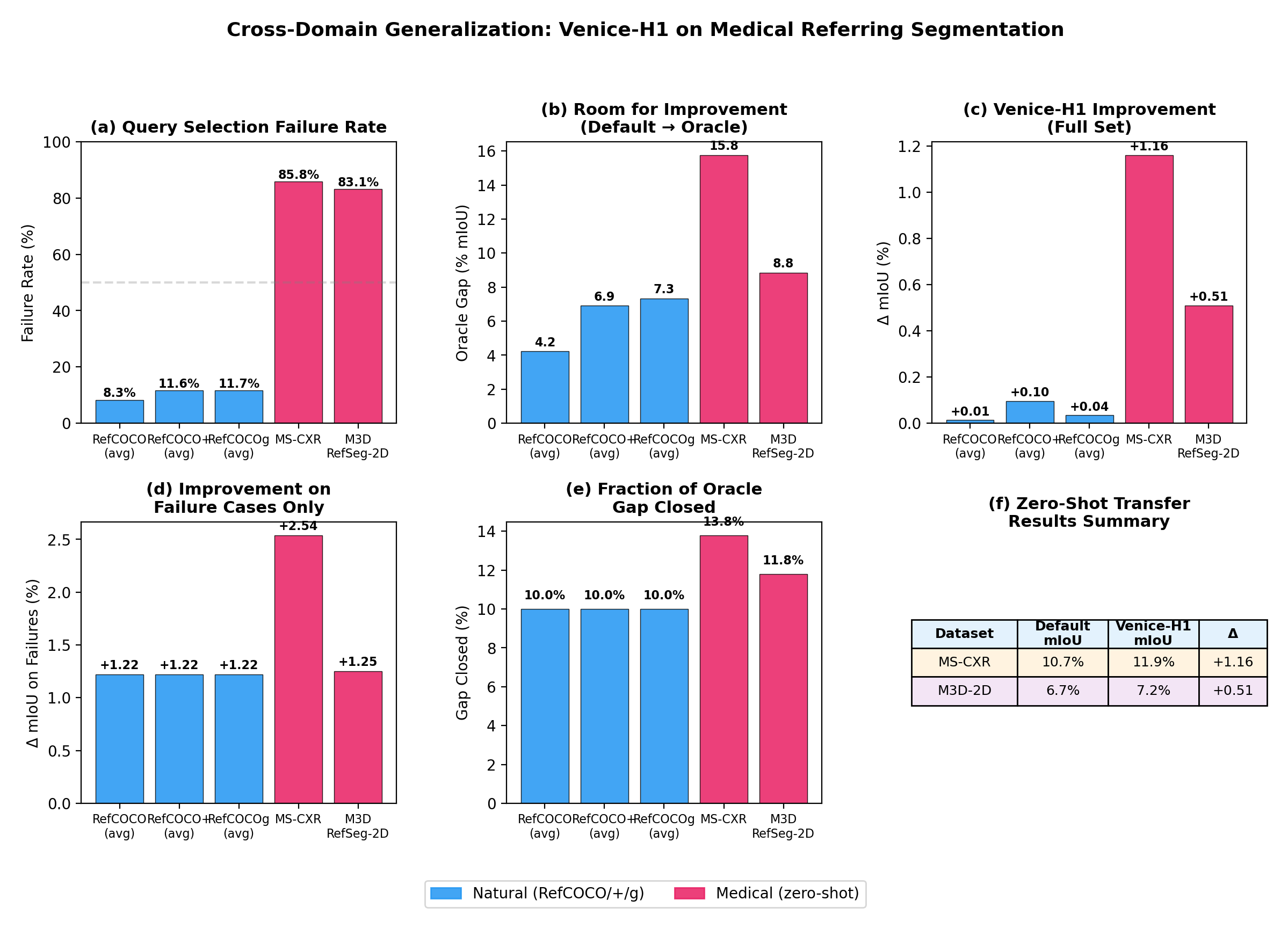}
\caption{\textbf{Cross-domain analysis.}
Natural images (blue) vs.\ medical datasets (pink) in a zero-shot setting.
(a)~Failure rates.
(b)~Best-query gaps.
(c--d)~\vH{} gains on both domains.}
\label{fig:supp_medical}
\end{figure*}

\paragraph{Cross-domain comparison.}
\Cref{fig:supp_medical} compares failure rates and best-query gaps between natural image benchmarks and zero-shot medical evaluation.
Medical data exhibits dramatically higher failure rates (83--86\% vs.\ 8--14\%) and larger best-query gaps (8--16\% vs.\ 3--8\%), yet \vH{} achieves positive gains on both domains.

\begin{figure*}[t]
\centering
\includegraphics[width=0.9\textwidth]{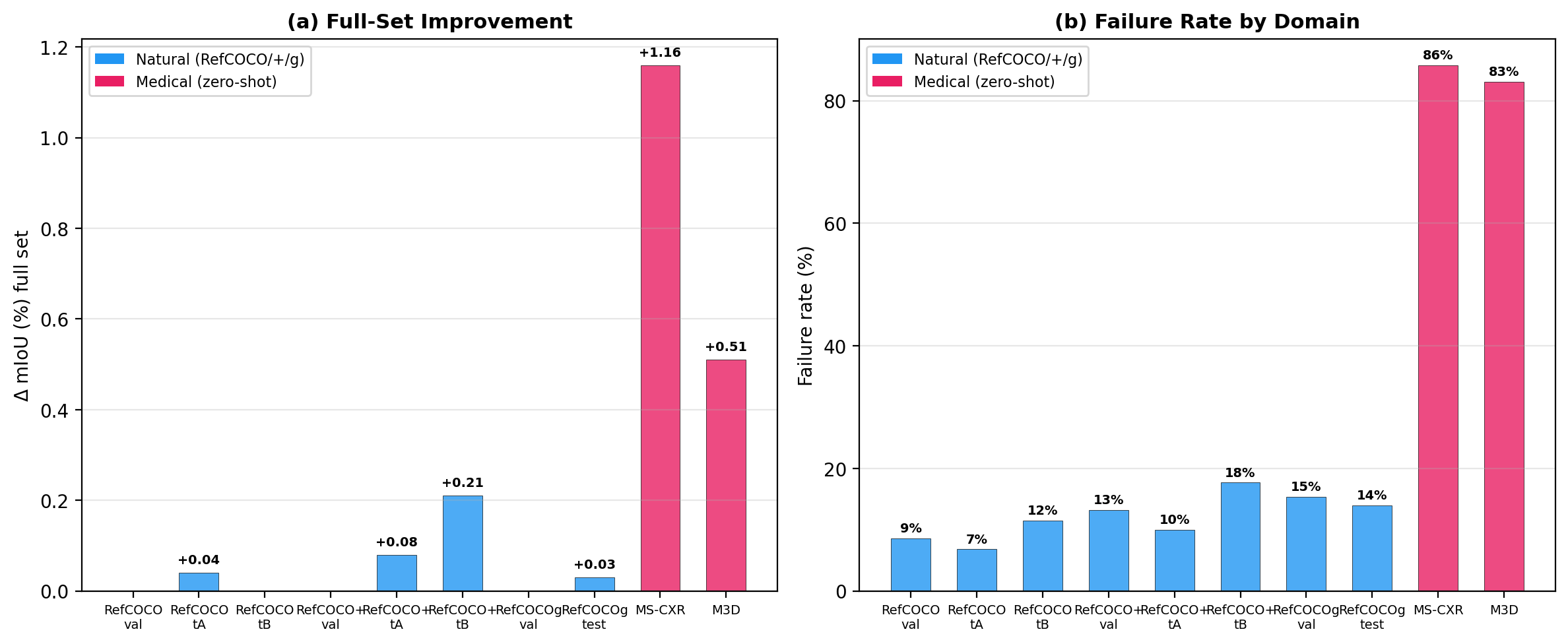}
\caption{\textbf{Unified view across all benchmarks.}
(a)~Full-set $\Delta$ mIoU: natural image splits (blue) and zero-shot medical splits (pink).
(b)~Failure rate by domain: medical data has ${\sim}10\times$ higher failure rates, creating more opportunities for re-ranking.}
\label{fig:combined}
\end{figure*}

\paragraph{Unified view.}
\Cref{fig:combined} provides a unified view of \vH{}'s improvements across all benchmarks.
On natural images, gains are modest but consistent (the baseline is near-optimal).
On medical data, the larger failure rates (83--86\%) yield proportionally larger absolute gains (+0.51--1.16\%), supporting the claim that \vH{}'s value increases when the upstream model struggles most.

\paragraph{Medical qualitative examples.}
\Cref{fig:med_qualitative} (in the Visual Gallery, \Cref{sec:galleries}) shows the top-3 improvements per medical dataset in a zero-shot setting.
Despite severe domain shift (natural images $\to$ chest X-rays and CT scans), \vH{} identifies suboptimal default selections and redirects to dramatically better alternatives, recovering +37--66\% IoU from near-zero baselines.

\paragraph{Per-sample IoU comparison.}
Per-sample IoU bar charts for six medical examples (\Cref{fig:med_iou_bars} in the Visual Gallery) clearly show how \vH{} narrows the gap between the default selection and the best-query upper bound.

\paragraph{Grid signatures on medical data.}
\Cref{fig:med_grids} (Visual Gallery) visualizes multi-scale grid signatures for medical samples, demonstrating that the spatial structure captured by grid signatures is meaningful even on out-of-distribution medical images---the re-ranked query exhibits more focused activation in the correct anatomical region.

\FloatBarrier
\section{Conclusion}
\label{sec:conclusion}
\balance

We introduced \vH{}, a lightweight failure-aware re-ranking framework for Referring Image Segmentation.
Starting from the observation that state-of-the-art multi-query models like DeRIS select a suboptimal query in 7--18\% of evaluation samples, we focused on these failure cases where a better alternative exists, with best-query gaps ranging from 3.2\% to 11.1\% mIoU across benchmarks.
Our approach introduces multi-scale grid signatures---compact 675-dimensional spatial descriptors obtained by pooling mask probabilities onto $4{\times}4$, $8{\times}8$, and $16{\times}16$ grids---that capture mask quality at three complementary resolutions.
A Transformer-based re-ranker with a Failure Gate (AUC 0.78--0.82 across all splits) and an IoU-regression gain predictor detects suboptimal selections and redirects to a better query, while a gated decision mechanism preserves baseline accuracy on the majority of samples.
Instantiated on DeRIS-L and DeRIS-B---two multi-query backbones with different capacities---\vH{} achieves strictly positive $\Delta_{\mathrm{fail}}$ on all 16/16 (split, backbone) pairs, with bootstrap 95\% CIs excluding zero, harmful-switch rates below 0.53\%, and gate AUC 0.76--0.80.
The threshold stability analysis ($\tau \in [0, 0.7]$: $\Delta_{\mathrm{fail}}$ constant in $[+1.408, +1.409]$), Top-$K$ monotonicity ($K{=}3$: $\Delta_{\mathrm{fail}} = +2.20$), and LODO generalization ($\Delta_{\mathrm{fail}} = +3.56$ on held-out splits) confirm that the selective behavior is robust and not an artifact of a particular operating point or training distribution.
Our ablation demonstrates that: (1)~multi-scale grid signatures outperform both scalar-only features and any single grid scale; (2)~IoU regression on all samples is critical when training data has low failure rates; and (3)~boundary energy provides consistent improvements.
The approach adds ${\sim}$11\,M trainable parameters, requires ${\sim}$3 minutes of training on cached features, and introduces $<1$\,ms inference overhead.

\paragraph{Limitations and future work.}
While the failure gate achieves strong AUC, the overall mIoU improvement is modest because failure samples constitute only 7--18\% of evaluation data, and closing the best-query gap requires not just detecting failures but also accurately selecting the right alternative query.
A fundamental bottleneck is the extreme class imbalance in training: only 0.28\% of training samples exhibit failures, starving the gate and gain predictor of positive examples.
Future work could explore contrastive and pairwise ranking losses, data augmentation for failure cases, end-to-end fine-tuning with the backbone, and extension to mask refinement beyond query selection.
Our zero-shot cross-domain experiments on medical referring segmentation benchmarks (MS-CXR, M3D-RefSeg-2D) show that \vH{}'s failure-detection and re-ranking signals transfer across domains: the model improves query selection by +1.16 and +0.51 mIoU points without any RIS-backbone fine-tuning, despite the severe distribution shift from natural to medical images.
Because \vH{} operates solely on candidate masks and cached scores, it is backbone-decoupled: any multi-query RIS (or dense-prediction) system that produces multiple hypotheses can plug in the module without backbone modification, suggesting a general failure-aware selection principle for multi-query pipelines.

\FloatBarrier

{\small
\bibliographystyle{plain}
\bibliography{references}

\begin{thebibliography}{10}

\bibitem{banino2018}
Andrea Banino, Caswell Barry, Benigno Uria, Charles Blundell, Timothy Lillicrap, Dharshan Kumaran, Eleanor Maguire, and Demis Hassabis.
\newblock Vector-based navigation using grid-like representations in artificial agents.
\newblock {\em Nature}, 557:429--433, 2018.
\newblock \href{https://doi.org/10.1038/s41586-018-0102-6}{doi:10.1038/s41586-018-0102-6}.

\bibitem{softnms}
Navaneeth Bodla, Bharat Singh, Rama Chellappa, and Larry~S. Davis.
\newblock Soft-{NMS} -- improving object detection with one line of code.
\newblock In {\em ICCV}, 2017.
\newblock \href{https://doi.org/10.1109/ICCV.2017.593}{doi:10.1109/ICCV.2017.593}.

\bibitem{mscxr}
Benedikt Boecking, Naoto Usuyama, Shruthi Bannur, Daniel~C. Castro, Anton Schwaighofer, Stephanie Hyland, Maria Wetscherek, Tristan Naumann, Aditya Nori, Javier Alvarez-Valle, Hoifung Poon, and Ozan Oktay.
\newblock {MS-CXR}: Making the most of text semantics to improve biomedical vision--language processing.
\newblock In {\em Computer Vision -- ECCV 2022}, LNCS vol.\ 13696, pages 1--21. Springer, 2022.
\newblock \href{https://doi.org/10.1007/978-3-031-20059-5_1}{doi:10.1007/978-3-031-20059-5\_1}.

\bibitem{mask2former}
Bowen Cheng, Ishan Misra, Alexander~G. Schwing, Alexander Kirillov, and Rohit Girdhar.
\newblock Masked-attention mask transformer for universal image segmentation.
\newblock In {\em CVPR}, 2022.
\newblock \href{https://doi.org/10.1109/CVPR52688.2022.00135}{doi:10.1109/CVPR52688.2022.00135}.

\bibitem{deris}
Ming Dai, Wenxuan Cheng, Jiang-jiang Liu, Sen Yang, Wenxiao Cai, Yanpeng Sun, and Wankou Yang.
\newblock {DeRIS}: Decoupling perception and cognition for enhanced referring image segmentation through loopback synergy.
\newblock {\em arXiv preprint arXiv:2507.01738}, 2025.
\newblock \href{https://doi.org/10.48550/arXiv.2507.01738}{arXiv:2507.01738}.

\bibitem{c3vg}
Ming Dai, Jian Li, Jiedong Zhuang, Xian Zhang, and Wankou Yang.
\newblock {C$^3$VG}: Multi-task visual grounding with coarse-to-fine consistency constraints.
\newblock In {\em Proceedings of the AAAI Conference on Artificial Intelligence (AAAI)}, 2025.
\newblock \href{https://doi.org/10.48550/arXiv.2501.06710}{arXiv:2501.06710}.

\bibitem{m3d}
Fan Bai, Yuxin Du, Tiejun Huang, Max Q.-H. Meng, and Bo Zhao.
\newblock {M3D}: Advancing 3{D} medical image analysis with multi-modal large language models.
\newblock {\em arXiv preprint arXiv:2404.00578}, 2024.
\newblock \href{https://doi.org/10.48550/arXiv.2404.00578}{doi:10.48550/arXiv.2404.00578}.

\bibitem{efn}
Guang Feng, Zhiwei Hu, Lihe Zhang, and Huchuan Lu.
\newblock Encoder fusion network with co-attention embedding for referring image segmentation.
\newblock In {\em CVPR}, 2021.
\newblock \href{https://doi.org/10.1109/CVPR46437.2021.01525}{doi:10.1109/CVPR46437.2021.01525}.

\bibitem{geifman2017selective}
Yonatan Geifman and Ran El-Yaniv.
\newblock Selective classification for deep neural networks.
\newblock In {\em NeurIPS}, 2017.
\newblock \href{https://proceedings.neurips.cc/paper/2017/hash/4a8423d5e91fda00bb7e46540e2b0cf1-Abstract.html}{NeurIPS Proceedings 2017 (Paper 7073)}.

\bibitem{selectivenet}
Yonatan Geifman and Ran El-Yaniv.
\newblock {SelectiveNet}: A deep neural network with an integrated reject option.
\newblock In {\em ICML}, 2019.
\newblock \href{https://proceedings.mlr.press/v97/geifman19a.html}{PMLR vol.\ 97}.

\bibitem{ssc}
Francisco Eiras, Kemal Oksuz, Adel Bibi, Philip H.~S. Torr, and Puneet~K. Dokania.
\newblock Segment, select, correct: A framework for weakly-supervised referring segmentation.
\newblock {\em arXiv preprint arXiv:2310.13479}, 2023.
\newblock Accepted to ECCV'24 Workshop (Instance-Level Recognition).
\newblock \href{https://doi.org/10.48550/arXiv.2310.13479}{doi:10.48550/arXiv.2310.13479}.

\bibitem{hafting2005microstructure}
Torkel Hafting, Marianne Fyhn, Sturla Molden, May-Britt Moser, and Edvard~I. Moser.
\newblock Microstructure of a spatial map in the entorhinal cortex.
\newblock {\em Nature}, 436:801--806, 2005.
\newblock \href{https://doi.org/10.1038/nature03721}{doi:10.1038/nature03721}.

\bibitem{sppnet}
Kaiming He, Xiangyu Zhang, Shaoqing Ren, and Jian Sun.
\newblock Spatial pyramid pooling in deep convolutional networks for visual recognition.
\newblock In {\em ECCV}, 2014.
\newblock \href{https://doi.org/10.1007/978-3-319-10578-9_23}{doi:10.1007/978-3-319-10578-9\_23}.

\bibitem{learningnms}
Jan Hosang, Rodrigo Benenson, and Bernt Schiele.
\newblock Learning non-maximum suppression.
\newblock In {\em CVPR}, 2017.
\newblock \href{https://doi.org/10.1109/CVPR.2017.701}{doi:10.1109/CVPR.2017.701}.

\bibitem{cmpc}
Shaofei Huang, Tianrui Hui, Si~Liu, Guanbin Li, Yunchao Wei, Jizhong Han, Luoqi Liu, and Bo~Li.
\newblock Referring image segmentation via cross-modal progressive comprehension.
\newblock In {\em CVPR}, 2020.
\newblock \href{https://doi.org/10.1109/CVPR42600.2020.01050}{doi:10.1109/CVPR42600.2020.01050}.

\bibitem{maskscoring}
Zhaojin Huang, Lichao Huang, Yongchao Gong, Chang Huang, and Xinggang Wang.
\newblock Mask scoring {R-CNN}.
\newblock In {\em CVPR}, 2019.
\newblock \href{https://doi.org/10.1109/CVPR.2019.00657}{doi:10.1109/CVPR.2019.00657}.

\bibitem{lscm}
Tianrui Hui, Si~Liu, Shaofei Huang, Guanbin Li, Sansi Yu, Faxi Zhang, and Jizhong Han.
\newblock Linguistic structure guided context modeling for referring image segmentation.
\newblock In {\em ECCV}, 2020.
\newblock \href{https://doi.org/10.1007/978-3-030-58545-7_4}{doi:10.1007/978-3-030-58545-7\_4}.

\bibitem{iounet}
Borui Jiang, Ruixuan Luo, Jiayuan Mao, Tete Xiao, and Yuning Jiang.
\newblock Acquisition of localization confidence for accurate object detection.
\newblock In {\em ECCV}, 2018.
\newblock \href{https://doi.org/10.1007/978-3-030-01264-9_48}{doi:10.1007/978-3-030-01264-9\_48}.

\bibitem{sam}
Alexander Kirillov, Eric Mintun, Nikhila Ravi, Hanzi Mao, Chloe Rolland, Laura Gustafson, Tete Xiao, Spencer Whitehead, Alexander~C. Berg, Wan-Yen Lo, Piotr Doll{\'a}r, and Ross Girshick.
\newblock Segment anything.
\newblock In {\em ICCV}, 2023.
\newblock \href{https://doi.org/10.1109/ICCV51070.2023.00371}{doi:10.1109/ICCV51070.2023.00371}.

\bibitem{lin2017focal}
Tsung-Yi Lin, Priya Goyal, Ross Girshick, Kaiming He, and Piotr Doll{\'a}r.
\newblock Focal loss for dense object detection.
\newblock In {\em ICCV}, 2017.
\newblock \href{https://doi.org/10.1109/ICCV.2017.324}{doi:10.1109/ICCV.2017.324}.

\bibitem{swinb}
Ze~Liu, Yutong Lin, Yue Cao, Han Hu, Yixuan Wei, Zheng Zhang, Stephen Lin, and Baining Guo.
\newblock Swin transformer: Hierarchical vision transformer using shifted windows.
\newblock In {\em ICCV}, 2021.
\newblock \href{https://doi.org/10.1109/ICCV48922.2021.00986}{doi:10.1109/ICCV48922.2021.00986}.

\bibitem{mai2020}
Gengchen Mai, Krzysztof Janowicz, Bo~Yan, Rui Zhu, Ling Cai, and Ni~Lao.
\newblock Multi-scale representation learning for spatial feature distributions using grid cells ({Space2Vec}).
\newblock In {\em ICLR}, 2020.
\newblock \href{https://openreview.net/forum?id=rJljdh4KDH}{OpenReview ICLR 2020}.

\bibitem{refcocog}
Junhua Mao, Jonathan Huang, Alexander Toshev, Oana Camburu, Alan Yuille, and Kevin Murphy.
\newblock Generation and comprehension of unambiguous object descriptions.
\newblock In {\em CVPR}, 2016.
\newblock \href{https://doi.org/10.1109/CVPR.2016.329}{doi:10.1109/CVPR.2016.329}.

\bibitem{gridcells}
Edvard~I. Moser, Yasser Roudi, Menno~P. Witter, Clifford Kentros, Tobias Bonhoeffer, and May-Britt Moser.
\newblock Grid cells and cortical representation.
\newblock {\em Nature Reviews Neuroscience}, 15(7):466--481, July 2014.
\newblock \href{https://doi.org/10.1038/nrn3766}{doi:10.1038/nrn3766}. PMID: 24917300.

\bibitem{clip}
Alec Radford, Jong~Wook Kim, Chris Hallacy, Aditya Ramesh, Gabriel Goh, Sandhini Agarwal, Girish Sastry, Amanda Askell, Pamela Mishkin, Jack Clark, Gretchen Krueger, and Ilya Sutskever.
\newblock Learning transferable visual models from natural language supervision.
\newblock In {\em ICML}, 2021.
\newblock \href{https://proceedings.mlr.press/v139/radford21a.html}{PMLR vol.\ 139}.

\bibitem{schaeffer2023}
Rylan Schaeffer, Mikail Khona, Tzuhsuan Ma, Crist\'{o}bal Eyzaguirre, Sanmi Koyejo, and Ila~R. Fiete.
\newblock Self-supervised learning of representations for space generates multi-modular grid cells.
\newblock In {\em NeurIPS}, 2023.
\newblock \href{https://proceedings.neurips.cc/paper_files/paper/2023/hash/schaeffer-grid-cells}{NeurIPS Proceedings 2023}.

\bibitem{beit3}
Wenhui Wang, Hangbo Bao, Li~Dong, Johan Bjorck, Zhiliang Peng, Qiang Liu, Kriti Aggarwal, Owais~Khan Mohammed, Saksham Singhal, Subhojit Som, and Furu Wei.
\newblock {BEiT}-3: Image as a foreign language: {BEiT} pretraining for vision and vision-language tasks.
\newblock In {\em CVPR}, 2023.
\newblock \href{https://doi.org/10.1109/CVPR56688.2023.01838}{doi:10.1109/CVPR56688.2023.01838}.

\bibitem{cris}
Zhaoqing Wang, Yu~Lu, Qiang Li, Xunqiang Tao, Yandong Guo, Mingming Gong, and Tongliang Liu.
\newblock {CRIS}: {CLIP}-driven referring image segmentation.
\newblock In {\em CVPR}, 2022.
\newblock \href{https://doi.org/10.1109/CVPR52688.2022.02101}{doi:10.1109/CVPR52688.2022.02101}.

\bibitem{oneref}
Linhui Xiao, Dunliang Kuang, Siyuan Huang, Shiguang Shan, and Xilin Chen.
\newblock {OneRef}: Unified one-tower expression grounding and segmentation with mask referring modeling.
\newblock In {\em NeurIPS}, 2024.
\newblock \href{https://proceedings.neurips.cc/paper_files/paper/2024}{NeurIPS Proceedings 2024}.

\bibitem{lavt}
Zhao Yang, Jiaqi Wang, Yansong Tang, Kai Chen, Hengshuang Zhao, and Philip H.~S. Torr.
\newblock {LAVT}: Language-aware vision transformer for referring image segmentation.
\newblock In {\em CVPR}, 2022.
\newblock \href{https://doi.org/10.1109/CVPR52688.2022.01738}{doi:10.1109/CVPR52688.2022.01738}.

\bibitem{refcoco}
Licheng Yu, Patrick Poirson, Shan Yang, Alexander~C. Berg, and Tamara~L. Berg.
\newblock Modeling context in referring expressions.
\newblock In {\em Computer Vision -- ECCV 2016}, LNCS vol.\ 9906, pages 69--84. Springer, 2016.
\newblock \href{https://doi.org/10.1007/978-3-319-46475-6_5}{doi:10.1007/978-3-319-46475-6\_5}.

\end{thebibliography}
}

\clearpage
\onecolumn

\appendix
\section{Extended Qualitative Results}
\label{sec:galleries}
\vspace{2pt}
\noindent
\begin{center}
\includegraphics[width=\textwidth,height=0.84\textheight,keepaspectratio]{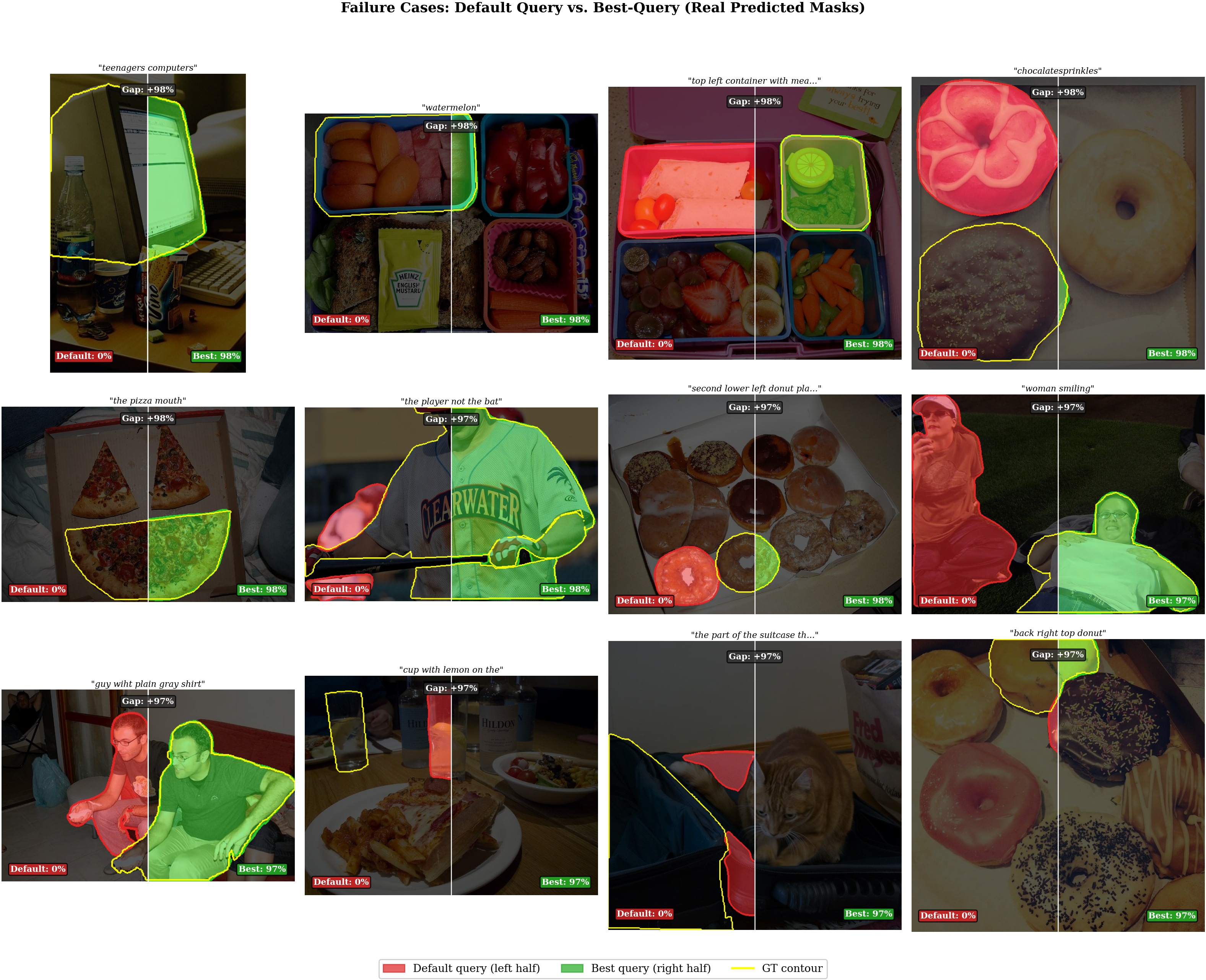}
\end{center}
\vspace{-6pt}
\captionof{figure}{Top-12 failure cases on RefCOCO val.
Each cell: default query (left, red overlay) vs.\ best query (right, green overlay).
Ground-truth contour in yellow; IoU gap at top.
A correct mask \emph{exists} among the candidates but is not selected by the default heuristic.}
\label{fig:failure_gallery}

\vspace{6pt}
\noindent
\begin{center}
\includegraphics[width=0.96\textwidth,height=0.88\textheight,keepaspectratio]{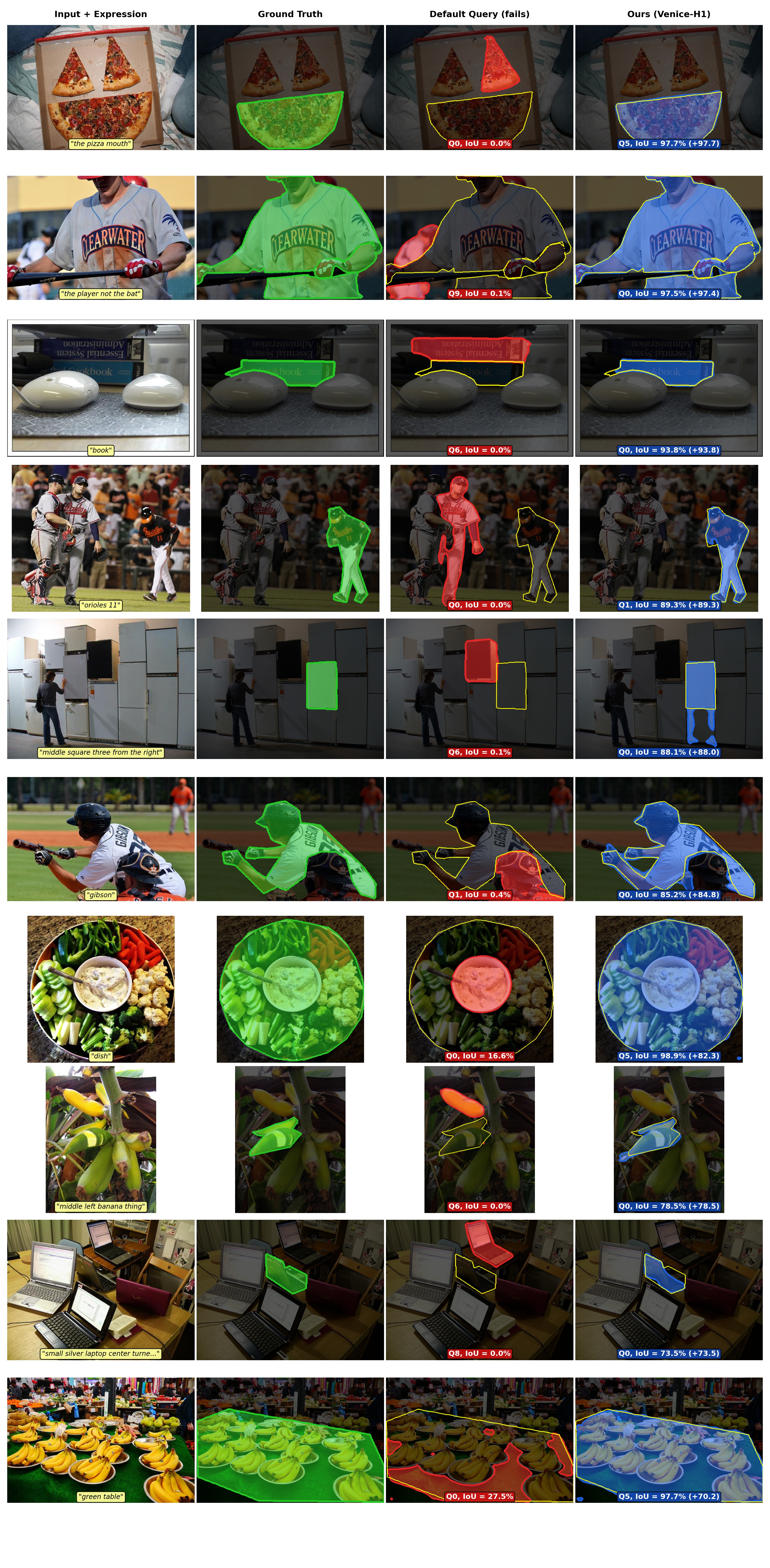}
\end{center}
\vspace{-8pt}
\captionof{figure}{Extended RefCOCO qualitative gallery (val, 10 additional examples).
Input + expression $\to$ ground truth (green) $\to$ default query (red, with IoU) $\to$ Venice-H1 (blue, with IoU and gain).}
\label{fig:refcoco_gallery}

\vspace{6pt}
\noindent
\begin{center}
\includegraphics[width=0.96\textwidth,height=0.42\textheight,keepaspectratio]{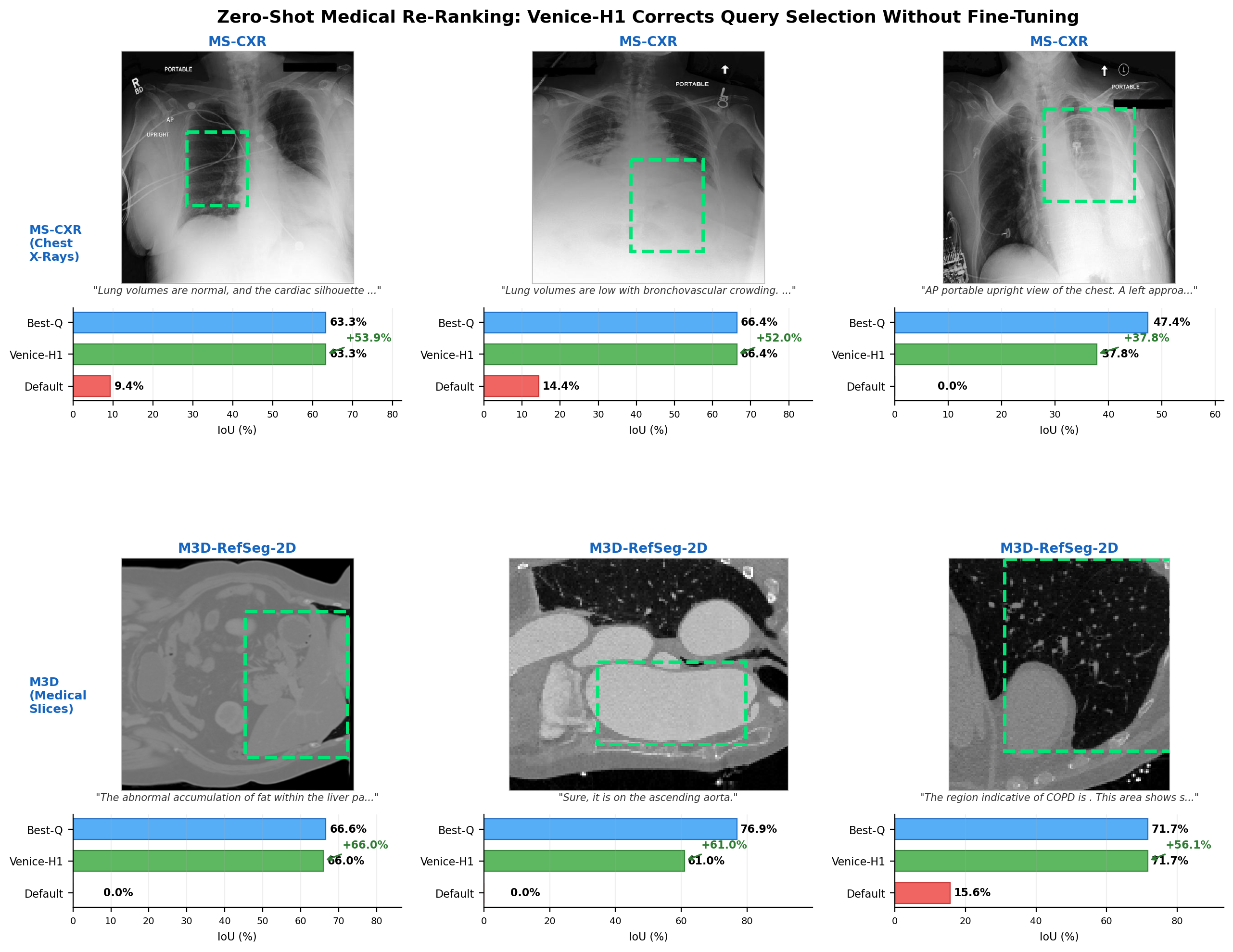}
\end{center}
\vspace{-8pt}
\captionof{figure}{Zero-shot medical re-ranking (no fine-tuning).
Top: MS-CXR chest X-rays; bottom: M3D-RefSeg-2D 3D medical slices.
Default (red), Venice-H1 re-ranked (green), best-query upper bound (blue).}
\label{fig:med_qualitative}

\vspace{4pt}
\noindent
\begin{minipage}[t]{0.54\textwidth}
\centering
\includegraphics[width=\textwidth]{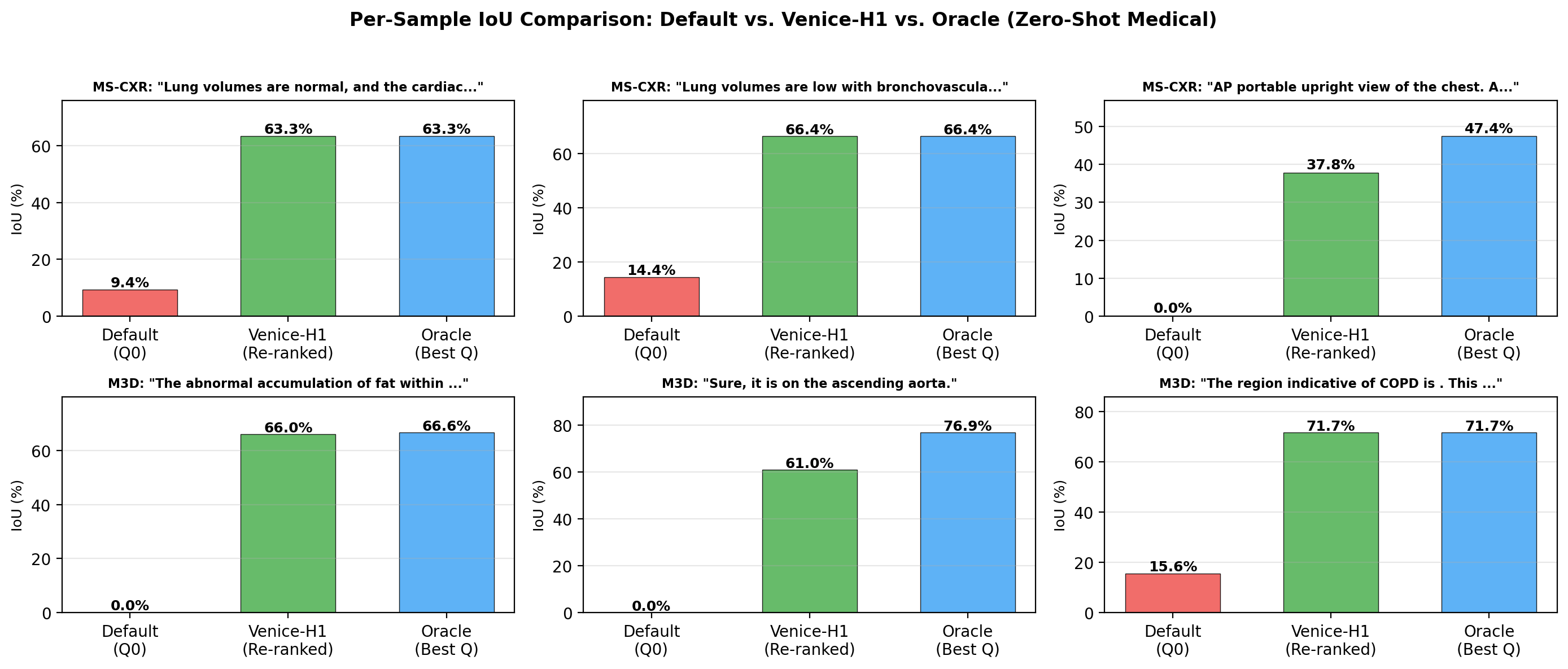}
\vspace{-6pt}
\captionof{figure}{Per-sample IoU comparison on medical data. Default (red), Venice-H1 (green), oracle (blue).}
\label{fig:med_iou_bars}
\end{minipage}%
\hfill
\begin{minipage}[t]{0.44\textwidth}
\centering
\includegraphics[width=\textwidth]{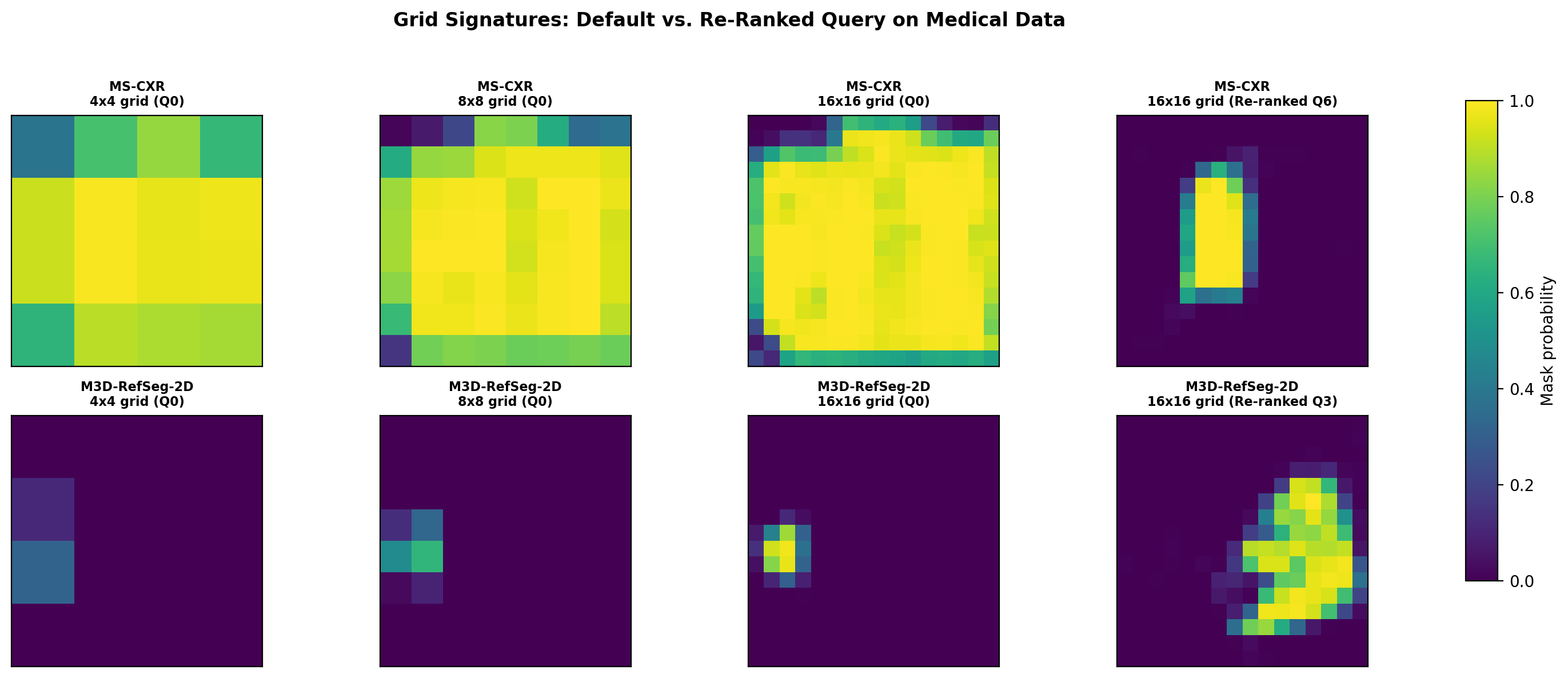}
\vspace{-6pt}
\captionof{figure}{Grid signatures on medical data. Multi-scale activations for Q0 (cols 1--3) and re-ranked query (col 4).}
\label{fig:med_grids}
\end{minipage}

\end{document}